\documentclass[sensors,perspective,accept,moreauthors]{Definitions/mdpi} 

\firstpage{1} 
\pubvolume{1}
\issuenum{1}
\articlenumber{0}
\pubyear{2026}
\copyrightyear{2026}
\externaleditor{Ben Hamza} 
\datereceived{07 May 2026} 
\daterevised{28 June 2026} 
\dateaccepted{15 July 2026} 
\datepublished{ }

\Title{Multi-Exposure HDR Imaging: A Review of Pixel-Level and Feature-Level Reconstruction Methods} 

\Author{Qian Tao $^{1}$\orcidA{}, Wei Wang $^{1,}$*\orcidB{}, Chaobing Zheng $^{2}$\orcidC{} and Zhengguo Li $^{3}$\orcidD{}} 

\AuthorNames{Qian Tao, Wei Wang, Chaobing Zheng and Zhengguo Li}

\address{%
$^{1}$ \quad School of Computer Science and Technology, Hubei Province Key Laboratory of Intelligent Information Processing and Real-Time Industrial System, Wuhan University of Science and Technology, Wuhan 430081, China; tqian@wust.edu.cn \\
$^{2}$ \quad School of Electronic Information, Wuhan University of Science and Technology, Wuhan 430081, China; zhengchaobing@wust.edu.cn \\
$^{3}$ \quad Institute of Advanced Intelligence and Computing, A*STAR, Singapore 138632; li\_zhengguo@a-star.edu.sg }

\corres{Correspondence: wangwei8@wust.edu.cn}

\abstract{Multi-exposure is an efficient way to capture real-world high-dynamic-range (HDR) scenes. However, HDR imaging suffers from severe ghosting artifacts in dynamic scenes due to the temporal gap between sequential exposures. In this article, we categorize the literature on two important topics on HDR imaging: multi-exposure fusion (MEF) and ghost removal. Conventional filter-based and data-driven methods are studied in pixel space and feature space. For popular deep learning-based approaches, we provide a granular taxonomy based on their alignment and fusion domains: pixel-space methods, which typically employ explicit motion compensation such as optical flow or spatial transformers, and feature-space methods, which leverage implicit alignment through deformable convolutions, attention mechanisms, or latent representation merging. Representative works are compared across different supervision settings, and key design principles are summarized. In addition, this survey summarizes commonly used datasets and evaluation metrics, discussing their applicability under diverse output forms. Finally, major bottlenecks and promising directions for future research are outlined.}
\begin{document}
\section{Introduction}
Natural scenes exhibit a wide luminance span, where intense sunlight and deep shadows often coexist. In contrast, the dynamic range of an image sensor is limited, so a single exposure can only capture a restricted interval of this span \cite{mertens2007exposure, yan2019attention}. As a result, highlights are easily saturated and washed out, shadows are prone to underexposure and clipping, and many scene details are lost during acquisition \cite{wu2022self,guo2023low,wu2023generation,wu2024overall}. To record scene content more faithfully, high-dynamic-range (HDR) imaging has become an important research direction. Existing HDR pipelines include hardware-level advances, multi-sensor imaging, and display-side processing such as tone mapping \cite{kim2023joint}. Among them, capturing a multi-exposure sequence is the most common and practical choice \cite{hu2013hdr}. It requires no hardware modification and only relies on multiple low-dynamic-range (LDR) images captured at different exposure levels, whose complementary information can cover a broader luminance range. This setting has therefore been widely adopted in mobile photography and \mbox{industrial vision}.

The core idea of multi-exposure fusion is straightforward \cite{mertens2007exposure}. A set of LDR images with different exposures is taken as input, and a single output image is produced with richer details and a more natural appearance that better matches human perception. Early methods mainly depended on hand-crafted fusion rules and multi-scale decompositions, selecting and combining details and contrast either in the spatial domain or in transform domains \cite{cai2018learning, chen2025ultrafusion}. With the progress of multi-scale representation learning and deep models, multi-exposure fusion has achieved clear improvements in detail preservation, local contrast, and structural stability, and the body of related work has grown rapidly.

In real-world capture, however, conditions are rarely ideal. Multi-exposure sequences often involve handheld camera shake and scene motion, leading to displacement and occlusion across exposures. Directly applying fusion rules then tends to introduce ghosting and structural breaks \cite{xiao2022deep}. At the same time, information loss under underexposure and overexposure can destabilize luminance hierarchy and color relationships, causing artifacts such as brightness reversal, color shift, and unnatural contrast. In other words, the challenges in multi-exposure imaging arise not only from limited dynamic range but also from misalignment in dynamic scenes and luminance and color drift induced by \mbox{exposure differences}.

To avoid conceptual confusion, this survey discusses multi-exposure fusion within a broader multi-exposure imaging framework. In terms of output form, the literature largely follows two formulations. One family directly produces a display-ready LDR fused image, aiming for a natural appearance and faithful detail rendition, and typically does not explicitly recover HDR radiance. The other family targets HDR reconstruction, estimating an HDR image or a linear HDR representation first and then applying tone mapping for visualization \cite{zhang2023self, yan2019attention}. Since both families share the same multi-exposure inputs and many papers also present tone-mapped results, terminology is often used in an overlapping manner. This survey therefore clarifies task definitions upfront and then reviews each method according to its primary objective and key design choices.

\begin{figure}[H]
    \includegraphics[width=\linewidth]{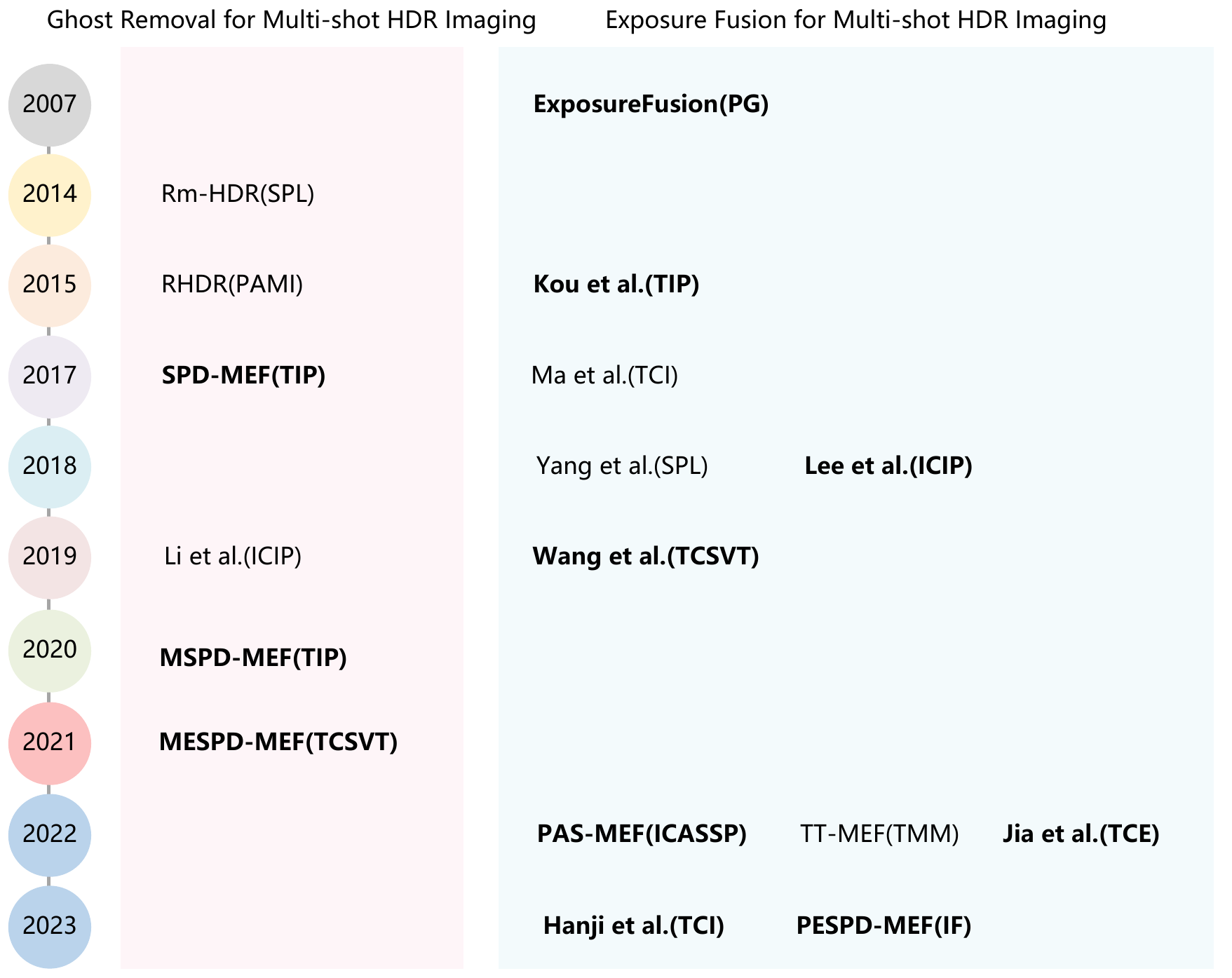}
    \caption{A concise milestone of pixel-space MEF methods. The timeline summarizes the evolution from hand-crafted exposure weighting and pyramid-based blending to structure-aware, edge-preserving, and reliability-aware fusion. Representative methods include Exposure Fusion~\cite{mertens2007exposure}, Rm-HDR~\cite{lee2014ghost}, RHDR~\cite{oh2014robust}, Kou et al.~\cite{kou2015gradient}, Ma et al.~\cite{ma2017multi}, Yang et al.~\cite{yang2018multi}, Lee et al.~\cite{lee2018multi}, SPD-MEF~\cite{ma2017robust}, Li et al.~\cite{li2019hybrid}, Wang et al.~\cite{wang2019detail}, MSPD-MEF~\cite{li2020fast}, MESPD-MEF~\cite{li2021detail}, PAS-MEF~\cite{karakaya2022pas}, TT-MEF~\cite{xu2021tensor}, Jia et al.~\cite{jia2022multi}, Hanji et al.~\cite{hanji2023robust}, and PESPD-MEF~\cite{zhang2023multi}.}
    \label{fig:tradition1}
\end{figure}

\begin{figure}[H] 
    \begin{adjustwidth}{-\extralength}{0cm} 
        \centering
        \includegraphics[width=\linewidth]{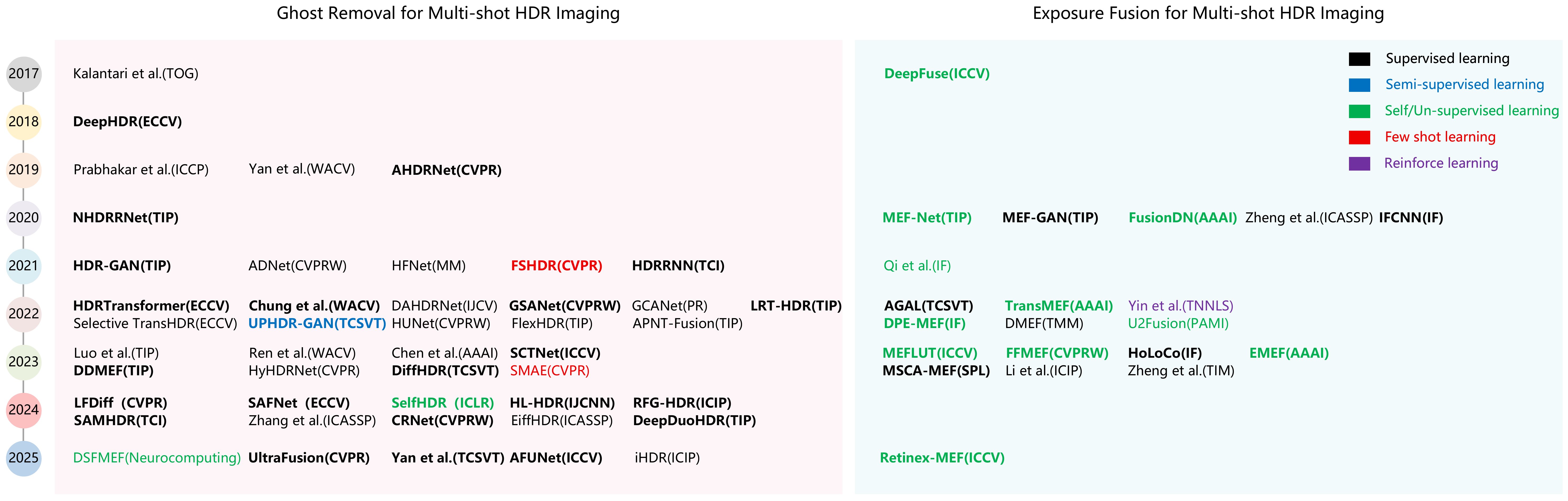} 
    \end{adjustwidth}
    \caption{A concise milestone of feature-space MEF methods. The timeline summarizes the evolution from CNN-based feature fusion to attention-based alignment, Transformer-based interaction, and generative restoration. Representative methods include Kalantari et al.~\cite{kalantari2017deep}, DeepHDR~\cite{wu2018deep}, Prabhakar et al.~\cite{prabhakar2019fast}, Yan et al.~\cite{yan2019multi}, AHDRNet~\cite{yan2019attention}, NHDRRNet~\cite{yan2020deep}, HDR-GAN~\cite{niu2021hdr}, ADNet~\cite{liu2021adnet}, HFNet~\cite{xiong2021hierarchical}, FSHDR~\cite{prabhakar2021labeled}, HDRRNN~\cite{prabhakar2021self}, HDRTransformer~\cite{liu2022ghost}, Chung et al.~\cite{chung2022high}, DAHDRNet~\cite{yan2022dual}, GSANet~\cite{li2022gamma}, GCANet~\cite{yan2022high}, LRT-HDR~\cite{mai2022deep}, Selective TransHDR~\cite{song2022selective}, UPHDR-GAN~\cite{li2022uphdr}, HUNet~\cite{yan2022lightweight}, FlexHDR~\cite{catley2022flexhdr}, APNT-Fusion~\cite{chen2022attention}, Luo et al.~\cite{luo2023multi}, Ren et al.~\cite{ren2023robust}, Chen et al.~\cite{chen2023improving}, SCTNet~\cite{tel2023alignment}, DDMEF~\cite{tan2023deep}, HyHDRNet~\cite{yan2023unified}, DiffHDR~\cite{yan2023toward}, SMAE~\cite{yan2023smae}, LFDiff~\cite{hu2024generating}, SAMHDR~\cite{li2024single}, SAFNet~\cite{kong2024safnet}, Zhang et al.~\cite{zhang2024efficient}, SelfHDR~\cite{zhang2023self}, CRNet~\cite{yang2024crnet}, HL-HDR~\cite{zhang2024hl}, EiffHDR~\cite{zhang2024eiffhdr}, RFG-HDR~\cite{lee2024rfg}, DeepDuoHDR~\cite{alpay2024deepduohdr}, DSFMEF~\cite{zhao2025single}, UltraFusion~\cite{chen2025ultrafusion}, Yan et al.~\cite{yan2024dynamic}, AFUNet~\cite{li2025afunet}, iHDR~\cite{yuan2025ihdr}, DeepFuse~\cite{ram2017deepfuse}, MEF-Net~\cite{1ma2019}, MEF-GAN~\cite{xu2020mef}, FusionDN~\cite{xu2020fusiondn}, Zheng et al.~\cite{zheng2020exposure}, IFCNN~\cite{zhang2020ifcnn}, Qi et al.~\cite{qi2021deep}, AGAL~\cite{liu2022attention}, TransMEF~\cite{qu2022transmef}, Yin et al.~\cite{yin2021automatic}, DPE-MEF~\cite{han2022multi}, DMEF~\cite{wu2022dmef}, U2Fusion~\cite{xu2020u2fusion}, MEFLUT~\cite{jiang2023meflut}, FFMEF~\cite{zheng2023efficient}, HoLoCo~\cite{liu2023holoco}, EMEF~\cite{liu2023emef}, MSCA-MEF~\cite{liu2023multi}, Li et al.~\cite{li2023neural}, Zheng et al.~\cite{zheng2023neural}, and Retinex-MEF~\cite{bai2025retinex}.}
    \label{fig:deep1}
\end{figure}

From the perspective of the main challenges being addressed, we organize multi-exposure HDR imaging methods along two major lines. The first line concerns the fusion of aligned LDR images with different exposures. It focuses on detail loss and color drift induced by underexposure and overexposure \cite{zhu2026slcformer,he2025disentangle}, as well as bright-order-reversal (BOR) artifacts and the loss of global contrast (or scene depth), aiming to maintain a natural and reliable luminance hierarchy and color relationship after fusion. The second line focuses on ghost removal. It addresses structural inconsistency caused by cross-exposure motion and occlusion, with the goal of keeping fused results sharp and stable in dynamic regions. Although many methods involve both motion handling and exposure correction, we categorize each work by its primary objective as stated by the authors and supported by the core mechanism and dominant experimental evidence, while auxiliary components addressing the other issue do not change the primary route. Along these two lines, we further summarize representative approaches under traditional pipelines, as shown in Figure \ref{fig:tradition1}, and deep learning-based methods, as shown in Figure \ref{fig:deep1}, and discuss the mainstream strategies and key modules under different supervision settings. Figure~\ref{fig:tradition1} summarizes the main evolution route of pixel-space MEF methods. Early methods mainly rely on hand-crafted exposure measures, contrast measures, and multi-scale blending to select well-exposed pixels and suppress visible seams. Later methods further introduce edge-preserving filtering, detail enhancement, structural patch modeling, and consistency checking, so that fusion is no longer only guided by exposure quality but also by local structure and reliability. Therefore, the development of pixel-space methods can be understood as a gradual transition from simple pixel-level exposure selection to structure-aware and reliability-aware fusion. Figure~\ref{fig:deep1} summarizes the main evolution route of feature-space MEF methods. Compared with pixel-space methods, feature-space methods shift the fusion process from hand-crafted rules to learned representation modeling \cite{karadjuzovic2017assessment, peng2021two}. Early deep models mainly adopt CNN-based encoder--decoder networks for exposure correction and feature fusion. Later methods introduce attention mechanisms, deformable alignment, non-local interaction, recurrent memory, and Transformer-based global modeling to handle motion, saturation, and unreliable regions. More recent GAN- and diffusion-based methods further treat severely saturated, occluded, or misaligned regions as restoration or content completion problems, indicating a shift from deterministic fusion toward reliability-aware selection and generative HDR restoration \cite{zhang2025ghost, alotaibi2023quality, aslam2024qualitynet}. The reviewed papers were collected from IEEE Xplore, ACM Digital Library, SpringerLink, ScienceDirect/Elsevier, arXiv, and Google Scholar. The search keywords included ``HDR reconstruction'', ``multi-exposure HDR'', ``multi-exposure fusion'', ``exposure fusion'', ``ghost removal'', ``HDR deghosting'', and ``dynamic HDR imaging'' \cite{fang2017perceptual, russell2024asymmetric}. This survey mainly includes representative works published from 2007 to the present that are directly related to multi-exposure fusion, HDR reconstruction, or deghosting, while papers without clear relevance to multi-exposure HDR imaging were excluded \cite{yoon2022multi, qi2020precise}.
For clarity and consistency, Table~\ref{tab:abbreviations} summarizes the abbreviations used throughout this survey and provides their corresponding full names and brief explanations.

\vspace{-6pt}
\begin{table}[H]
    \renewcommand{\arraystretch}{1.3}
    \caption{List of abbreviations used in this paper.} 
    \label{tab:abbreviations}
    \begin{adjustwidth}{-\extralength}{0cm}
        \centering
        \footnotesize
        \setlength{\tabcolsep}{3pt}
        \resizebox{\linewidth}{!}{
            \begin{tabular}{clclcl}
            \toprule
            \textbf{Abbr.} & \textbf{Full Name} &
            \textbf{Abbr.} & \textbf{Full Name} &
            \textbf{Abbr.} & \textbf{Full Name} \\
            \midrule
            HDR & High Dynamic Range &
            LDR & Low Dynamic Range &
            MEF & Multi-exposure Fusion \\
            BOR & Brightness-Order Reversal &
            CRF & Camera Response Function &
            SNR & Signal-to-Noise Ratio \\
            ISO & International Organization for Standardization &
            ISP & Image Signal Processor &
            CMOS & Complementary Metal--Oxide--Semiconductor \\
            AE & Auto-Exposure &
            LER & Large Exposure Ratio &
            EPS & Edge-Preserving Smoothing \\
            WGIF & Weighted Guided Image Filter &
            GGIF & Gradient Domain Guided Image Filter &
            GFU & Guided Filtering for Upsampling \\
            CNN & Convolutional Neural Network &
            GAN & Generative Adversarial Network &
            LUT & Lookup Table \\
            RNN & Recurrent Neural Network &
            GT & Ground Truth &
            SOTA & State of the Art \\
            PSNR & Peak Signal-to-Noise Ratio &
            MSE & Mean Squared Error &
            SSIM & Structural Similarity Index Measure \\
            MS-SSIM/MSSSIM & Multi-Scale Structural Similarity Index Measure &
            HDR-VDP-2 & High-Dynamic-Range Visual Difference Predictor 2 &
            MEF-SSIM & Multi-exposure Fusion Structural Similarity \\
            MI & Mutual Information &
            FMI & Feature Mutual Information &
            NMI & Normalized Mutual Information \\
            QNCIE & Nonlinear Correlation Information Entropy &
            SD & Standard Deviation &
            He & Entropy \\
            CC & Correlation Coefficient &
            AG & Average Gradient &
            VIF & Visual Information Fidelity \\
            TMQI & Tone-Mapped Image Quality Index &
            QAB/F & Edge Preservation Metric &
            CE & Cross Entropy \\
            QP & Phase Congruency-based Fusion Metric &
            QW & Wang's Fusion Metric &
            QCB & Chen--Blum Fusion Metric \\
            QCV & Chen-Varshney Fusion Metric &
            NIQE & Natural Image Quality Evaluator &
            T & Training Set \\
            Te & Test Set &
            val & Validation Set &
            ~ & ~ \\
            \bottomrule
            \end{tabular}
        }
    \end{adjustwidth}
\end{table}

\section{Differently Exposed LDR Images by Multiple Shots}

An LDR image can be a raw image or an sRGB image. Let $\Phi$ be the radiant power that each pixel measures for a real-world HDR scene, i.e., the light that it collects. $\Phi$ can be thought of as the scene brightness. A raw image $I_i$ is captured by the $i$th-shot as \cite{1Hasinoffsw2010}
\begin{align}
\label{imagingmodel}
I_i(p)=\min\{\frac{\Phi(p)\Delta t_i}{g_i}+I_0+e_i(p), I_{max}\},
\end{align}
where $p$ is a pixel, $\Delta t_i$ is the exposure time, and $g_i$ is the $i$th {sensor} gain. $I_0$ is a  constant offset representing the black point. $I_{max}$ is the saturation level, i.e., the maximum sensor value that can be recorded.  $e_i$ is the signal- and gain-dependent sensor noise. 

The corresponding sRGB image is denoted as $Z_i$ and is captured by 
\begin{align}
\label{imagingmodel2}
Z_{i,c}(p)=f_c(I_i(p))\; ;\; c\in \{R, G, B\},
\end{align}
where $f_c(\cdot)$ is a camera response function (CRF). The signal-to-noise-ratio (SNR) of the pixel $I_i(p)$ is \cite{1Hasinoffsw2010}
\begin{align}
\label{SNR}
SNR(I_i(p))=\frac{\Phi^2(p)\Delta t_i^2\cdot [I_i(p)<I_{max}]}{\Phi(p)\Delta t_i+\sigma^2_{read}+\sigma^2_{ADC}g_i^2},
\end{align}
where $[I_i(p)<I_{max}]$ indicates whether the pixel $I_i(p)$ is saturated. $\sigma^2_{read}$ and $\sigma^2_{ADC}$ denote the variances of the read-out noise and the analog-to-digital conversion noise, respectively; equivalently, $\sigma_{read}$ and $\sigma_{ADC}$ are their corresponding standard deviations. It can be easily verified that $\frac{\partial SNR(I_i(p))}{\partial \Delta t_i}>0$ and $\frac{\partial SNR(I_i(p))}{\partial g_i}<0$ when the pixel $I_i(p)$ is not saturated.

Similarly to \cite{11lizg2025}, three sets are defined as
 \begin{align}
 \Omega &=\{Z_1,\cdots, Z_K\},\\
 \Omega_f &=\{Z_{f_1}, \cdots, Z_{f_{\theta_1}}\}, \\
 \Omega_m &=\{Z_{m_1},\cdots, Z_{m_{\theta_2}}\},
 \end{align}
where $\Omega_f$ defines the set of sRGB images to be fused. $\theta_1$ is usually 3 and it can also be selected as 2. $\Omega_m$ determines the set of sRGB images for the computation of loss functions. The relationship among the three sets is
 \begin{align}
 \Omega_f\subseteq \Omega_m\subseteq \Omega.
 \end{align}
 
It should be noted that the sets $\Omega_f$ and $\Omega_m$ are the same for most existing MEF algorithms. 
All differently exposed sRGB images in the set $\Omega$ cover the whole dynamic range of the real-world HDR scene. Two popular methods to capture the images in the set $\Omega$ are: (1) fix the sensor gain while the exposure time is changed and (2) fix the exposure time while the sensor gain is changed \cite{Debevec97}. The former is more popular than the latter. Without loss of generality, it is assumed that
\begin{align}
\label{8}
\frac{\Delta t_1}{g_1}<\frac{\Delta t_2}{g_2}<\cdots <\frac{\Delta t_k}{g_k}. 
\end{align}

There are two different ways to combine the images in the set $\Omega_f$ together.   One is to estimate  the CRFs, convert all input  images into their corresponding HDR images, and merge all HDR images into one high-quality HDR image using weighted frame averaging~\cite{Debevec97}. The HDR image is converted into an 8-bit image by a tone mapping algorithm for display \cite{1farbman2008,1zhengjh2014,1vinker2021, Zhang2023}. The other is to directly fuse all input images into an information-enriched 8-bit image $Z_F$ using an exposure fusion algorithm. This submission focuses on the latter. The images in the set $\Omega_f$ are usually captured using a hand-held device. There might be camera movements or  moving objects in the set $\Omega_f$. There are ghosting artifacts in the fused image $Z_f$ when all images in the set $\Omega_f$ include moving objects and they are directly fused. Therefore, existing works on ghost removal are also studied in this paper.

\section{Exposure Fusion for Multi-Shot HDR Imaging}

The problem addressed by existing MEF algorithms is "How to preserve scene depth and fine details of the real-world HDR scene that are captured by the set $\Omega_f$ in the fused image $Z_F$ with neither halo nor BOR artifacts?". The existing MEF algorithms can be divided into fusion in pixel space and fusion in feature space. 

\subsection{MEF in Pixel Space}

It can be shown from Equations (\ref{imagingmodel})  and  (\ref{imagingmodel2}) that each image $Z_i$ in the set $\Omega_f$ is a noisy observation of the fused image $Z_F$. Based on this observation, an MEF algorithm in pixel space conducts a weighted sum of image pixels in the set $\Omega_f$ to obtain the fused image $Z_f$. 

The weights can be computed using filter-based methods and learned using data-driven methods. Brightness inversion and exposure inconsistency often arise from saturation, underexposure, nonlinear camera response, and local contrast variations. Even when images are well aligned, these factors may lead to unnatural brightness transitions, detail distortion, or color shifts \cite{xu2021tensor}. In pixel-space methods, such issues are typically handled by carefully designing weight maps and stabilizing the fusion process through spatial weighting, intensity normalization, or multi-scale reconstruction.
A standard pipeline first assesses exposure quality and local reliability across the input images, followed by the construction of spatially varying weight maps to guide fusion. The images are then combined through a multi-scale representation \cite{hanji2023robust}, commonly implemented via pyramid-based blending, to ensure smooth transitions across regions. Finally, additional refinement steps, such as exposure compensation or tone adjustment, are applied when necessary to enhance overall visual consistency.

Mertens et al. \cite{mertens2007exposure} first used contrast, saturation, and exposure to define the weights for all pixels in the set $\Omega_f$ and then fused all input images in pixel space to create an information-enriched 8-bit image using the Gaussian and Laplacian pyramids \cite{1Burt1983}. The depth of the real-world scene is well preserved in the fused image by the MEF algorithm in~\cite{mertens2007exposure}. However, it has a fundamental difficulty in preserving details in the highlight/shadow regions of the real-world HDR scene, although they are captured by the set $\Omega_f$. To address this issue, edge-preserving smoothing (EPS) pyramids were proposed in \cite{1lizg2017,1kou2017,wang2019detail,jia2022multi} using a weighted guided image filter (WGIF) \cite{16lizg2015} and gradient domain guided image filter (GGIF)~\cite{kou2015gradient}. Guided filtering for up-sampling (GFU) on top of the WGIF \cite{16lizg2015} was adopted to simplify the MEF algorithm in \cite{1lizg2017}. Two coefficients of the WGIF are only computed at two levels of the pyramids and they are up-sampled to obtain the coefficients at other levels. Weight maps are then computed using the gray components and the coefficients of the WGIF at all other levels. Since the EPS pyramids can smoothen the weights, the levels of the pyramids can be reduced. As such, the details in the highlight/shadow regions of the real-world HDR scene can be well preserved in the fused image if they are captured by the set $\Omega_f$. However, halo artifacts could be an issue for the algorithms in \cite{1lizg2017,1kou2017,wang2019detail,jia2022multi}. One more issue for the filter-based MEF algorithms is that fine details are lost. Thus, detail enhancement components were proposed in \cite{11lizg2012,1lizg2017,wang2019detail} to first extract fine details from all images in the set $\Omega_f$ and then add the fine details to the fused image.

An alternative method is to strengthen weight estimation. Lee et al. \cite{lee2018multi} makes the weights depend on which exposure adds more useful content relative to the others, rather than judging each exposure in isolation. PAS-MEF \cite{karakaya2022pas} further enriches the weighting cues by combining global structure information with luminance-dependent scoring and visual importance guidance, which helps to preserve details in both bright and dark areas without making the fusion unstable.
 Ma et al. \cite{ma2017multi} directly updates the fused image by optimizing a structure-driven quality objective so the output is gradually pushed toward better structural appearance. TT-MEF \cite{xu2021tensor} performs fusion in a transformed representation and separates coarse luminance trends from fine details, then fuses them with different rules so that global exposure looks balanced while textures are retained.
PESPD-MEF \cite{zhang2023multi} follows a decomposition strategy for extreme exposure cases. It explicitly boosts missing details in severely underexposed or overexposed regions and then fuses different components with different criteria before multi-scale reconstruction, which improves both perceived richness and overall consistency.  Unsupervised-learning-based MEF algorithms in \cite{1ma2019,jiang2023meflut} are also based on the GFU while  the WGIF is replaced by a GIF \cite{1he2013}. Instead of computing the weight maps using the full-size images  \cite{mertens2007exposure,1lizg2017,1kou2017}, the weight maps are learned from the down-sampled images and then up-sampled to the full-size using the GFU. All input images are finally fused  together with the up-sampled weight maps.

Overall, pixel-space MEF methods evolve around improving the reliability of exposure selection and the stability of multi-scale fusion. They do not have any constraint on the number of images in the set $\Omega_f$. By refining weight estimation, enhancing luminance consistency, and designing more robust reconstruction strategies, these methods progressively improve detail preservation, color fidelity, and robustness under extreme exposure conditions. However,  the BOR artifacts are produced by all the above filter-based MEF algorithms and two learning-based MEF algorithms in  \cite{1ma2019,jiang2023meflut} when two LER images are fused. This problem can be addressed by  applying exposure interpolation, intensity mapping, or response normalization to bring input images to a more consistent brightness domain.  Yang et al. \cite{yang2018multi} synthesizes intermediate exposures through intensity mapping functions (IMFs)~\cite{1Grossberg2003,16zhengcb2025} to handle large exposure gaps, while Hanji et al. \cite{hanji2023robust} estimates global exposure ratios using noise-aware correspondence modeling for more robust calibration.

\subsection{MEF in Feature Space}

Feature space usually provides 
more information than pixel space, whose inputs are usually two large-exposure-ratio (LER) images or three {consecutive} images. As most feature-based methods, deep learning-based luminance-consistent multi-exposure fusion aims to recover missing details from underexposed and overexposed inputs while maintaining stable luminance and color appearance. Most methods adopt multi-branch encoder--decoder architectures, where each exposure is first mapped into a feature representation. Cross-exposure features are then aligned and fused in a multi-scale manner \cite{zheng2020exposure}, followed by a reconstruction head that produces the final fused image or HDR output.

With supervision, these networks learn to restore correct luminance distributions and recover lost details under extreme exposure conditions. Training typically relies on pixel-level, structural, and perceptual objectives, sometimes guided by tone-mapped references. From the perspective of feature-space modeling, existing methods can be broadly categorized into three groups. CNN-based approaches mainly focus on feature normalization and correction before fusion. Attention-based multi-scale methods dynamically select and aggregate informative regions across exposures and scales, often enhanced by adversarial or contrastive objectives \cite{li2023neural}. Transformer-based methods further capture global dependencies, enabling long-range exposure interaction and more coherent \mbox{luminance modeling}.

\subsubsection{CNN-Based Methods}

CNN-based methods in the feature space mainly address the challenge of exposure-induced feature inconsistency. Under large exposure variations, reliable information is unevenly distributed across inputs, and direct feature fusion may lead to luminance shifts and color distortion. To mitigate this issue, these methods learn to map per-exposure features into a more consistent latent representation before fusion and decoding. This is typically achieved through conditional modulation, exposure-aware encoding, or structured correction modules.

Zheng et al. \cite{zheng2020exposure} adopt a feature-space pre-correction strategy. They first estimate a reliable mid-exposure representation using a physics-based exposure model, then train the network to predict residuals that compensate for luminance inconsistency and missing details, thus improving exposure continuity, reducing data dependency and alleviating luminance reversal. DMEF \cite{wu2022dmef} introduces Retinex-inspired feature-domain decomposition to separate illumination and reflectance. It conducts luminance alignment in the illumination space and texture fusion in the reflectance space, which preserves fine details under extreme exposures and stabilizes latent exposure relationships for more \mbox{reliable reconstruction}. 

\subsubsection{Attention and Transformer-Based Methods}

Methods in this category operate in the feature space and primarily focus on learning how to selectively aggregate information during fusion, i.e., determining which exposure should be trusted at each spatial location. In multi-exposure settings, underexposed inputs tend to provide more reliable details in dark regions, while overexposed inputs are more informative in bright regions. However, the distribution of reliable information varies significantly across scenes and spatial scales. To address this issue, attention mechanisms are introduced to model spatial and channel-wise dependencies, effectively learning adaptive fusion weight maps \cite{li2023neural}. Multi-scale architectures further complement this design by jointly capturing global luminance structure and local texture details. A typical pipeline extracts hierarchical or pyramid features, performs exposure-aware selection and fusion in the latent space, and refines the fused representation for final reconstruction, which helps to suppress luminance inconsistency and improve perceptual naturalness.

Following this idea, Li et al. \cite{li2023neural} introduce attention into exposure interpolation before fusion. For two-exposure inputs with a large exposure gap, they first reconstruct an intermediate exposure to improve luminance continuity, then feed the enhanced inputs into the fusion network for more stable aggregation. MEF-GAN \cite{xu2020mef} incorporates self-attention into an end-to-end fusion framework and combines adversarial learning with gradient-based constraints to alleviate the limitations of purely pixel-level supervision, leading to improved luminance consistency and more realistic textures. AGAL \cite{liu2022attention} further employs hierarchical attention mechanisms and introduces both global and local adversarial constraints to stabilize overall exposure distribution and local detail representation, while enhancing edge preservation during refinement, thereby reducing color distortion and texture blur under extreme exposure variations. HoLoCo \cite{liu2023holoco} instead leverages holistic and local contrastive constraints to guide more reliable feature aggregation, together with a color correction module that improves luminance and chrominance stability in challenging regions. Zheng et al. \cite{zheng2023neural} extend this line to a single-image setting, where a single LDR input is enhanced via an exposure-aware guidance branch to recover saturated regions, followed by multi-scale fusion for final reconstruction, improving applicability beyond multi-frame inputs. Additionally, IFCNN \cite{zhang2020ifcnn} serves as an early general-purpose fusion baseline, extracting features using convolutional layers and aggregating them via simple element-wise operations such as mean, max, or sum, without explicit attention modeling.

In MSCA-MEF \cite{liu2023multi}, MSCA-Net integrates both CNN and Transformer components to jointly capture local textures and global contextual dependencies. It further enhances cross-scale feature interaction through multi-scale feature fusion modules and atrous spatial pyramid pooling, achieving improved luminance consistency while preserving fine-\mbox{grained details}.

Overall, this branch formulates fusion as a learnable feature selection and aggregation process. Attention mechanisms determine where information should be drawn from, multi-scale modeling balances global structure with local details, and adversarial or contrastive objectives further improve luminance consistency and visual fidelity.

\subsubsection{Other Learning-Based Methods}

This category encompasses learning-based approaches that do not rely on standard CNN, attention, or Transformer architectures, but instead explore alternative learning paradigms such as no-reference supervision, physical or consistency-driven constraints, and model-free or preference-based fusion strategies \cite{yin2021automatic}.

This line of work operates in the feature space and focuses on modeling exposure relationships at a more global level. By leveraging long-range dependencies \cite{ram2017deepfuse}, these methods improve the consistency of luminance hierarchies across distant regions, which is particularly beneficial under large illumination variations and unstable global contrast. A typical pipeline converts multi-exposure inputs into interactive token or latent representations, learns exposure mappings and fusion strategies through global interaction, and then applies a decoder for local refinement and reconstruction. Compared with purely CNN-based designs, these approaches are generally more effective at preserving global luminance coherence and structural consistency, although they often require more careful design of tokenization, computational efficiency, and training stability.

When ground-truth supervision is not available, these methods shift the learning objective from direct output supervision to indirect constraints on the learning process. Instead of relying on paired references, they adopt no-reference quality measures or structure-aware constraints to guide exposure selection and fusion behavior. Alternatively, they enforce physical or statistical consistency by projecting results across exposure domains or decomposing images into interpretable components, with reconstruction consistency used to maintain stable luminance and color relationships and reduce artifacts.

From the perspective of quality-driven learning, many methods formulate fusion as a learnable weighted selection problem in the feature space. MEF-Net \cite{1ma2019} learns low-resolution weight maps and uses differentiable guided filtering for stable upsampling, followed by weighted fusion. It is trained with no-reference objectives such as MEF-SSIM, achieving a balance between efficiency and quality.  MEFLUT \cite{jiang2023meflut} improves deployment efficiency by learning a mapping from luminance to fusion weights and compressing it into a lightweight lookup table, enabling fast inference through table lookup and weighted aggregation. DeepFuse \cite{ram2017deepfuse} follows an earlier two-stream design that extracts features robust to extreme exposure differences and performs feature fusion and reconstruction without relying on explicit ground-truth supervision. Qi et al. \cite{qi2021deep} further improve no-reference learning by combining color-aware structural similarity with gradient consistency constraints, which helps to reduce edge blur and halo artifacts. FusionDN \cite{xu2020fusiondn} and U2Fusion~\cite{xu2020u2fusion} enhance selection mechanisms by estimating region-wise quality or informativeness to guide fusion, and introduce elastic weight consolidation to mitigate catastrophic forgetting in sequential learning. FFMEF \cite{zheng2023efficient} further explores an alternative formulation by predicting spatially adaptive filters instead of explicit weight maps, and stabilizes training using gradient-based unsupervised constraints.

From the perspective of consistency-based and interpretable modeling, the emphasis shifts to explicitly constraining the relationship between inputs and outputs. DPE-MEF~\cite{han2022multi} treats fusion as a perceptual enhancement process, using local exposure optimization to recover missing details and improve color naturalness. TransMEF \cite{qu2022transmef} adopts self-supervised multi-task learning, where synthetic degradations are generated through gamma correction, frequency perturbation, and region shuffling to learn robust exposure priors for reconstruction and fusion. Retinex-MEF \cite{bai2025retinex} addresses extreme overexposure by decomposing images into illumination and reflectance components, introducing a glare term, and enforcing reconstruction consistency to stabilize reflectance across exposures and reduce color contamination. Yin et al. \cite{yin2021automatic} formulate fusion as a reinforcement learning problem, gradually adjusting exposure levels through intermediate predictions; although reference-guided, it is often discussed within non-fully supervised settings due to its weak supervision nature.

Finally, EMEF \cite{liu2023emef} formulates multi-exposure fusion as an adaptive preference optimization problem. It learns to imitate multiple fusion styles and selects the optimal combination at test time using no-reference quality metrics, demonstrating that effective fusion can be achieved without a single end-to-end supervised model.

It is noticeable that the fused image approaches the set of images to {be} fused in the set $\Omega_f$ by existing MEF algorithms. On the other hand, the fused image should approach the real-world HDR scene. This new research topic might be investigated by leveraging the conventional wisdom of inferring better through seeing more and the asymmetry between training and testing stages of data-driven methods. Fine details in the highlight and shadow regions are learned by the MEF framework and preserved in the fused image, although they are not captured by the set of images to be fused. 

\section{Ghost Removal for Multi-Shot HDR Imaging}

Existing MEF algorithms assume that all images to be fused from a real-world HDR scene are already well aligned. Unfortunately, this assumption is not always true. Moving objects could appear in the set of images to be fused, and there are ghosting artifacts in the fused image. Existing works on ghost removal are summarized in this section. 

\subsection{Ghost Removal in Pixel Space}

Pixel-space deghosting methods aim to suppress artifacts caused by inconsistencies across multi-exposure images by operating directly on image intensities and explicit spatial correspondences. Since images captured with different exposures often vary in structure, visibility, and local appearance, directly fusing them can easily lead to ghosting, motion blur, or structural distortions. A key challenge lies in the fact that exposure changes affect brightness and visibility, making it difficult to define stable consistency cues under large radiance variations. To address this, most methods design cross-exposure consistency measures to evaluate the reliability of each region and guide the fusion process \cite{mertens2007exposure, huang2018color}. Regions considered reliable are used to fully exploit complementary exposure information for detail recovery and smooth tone transitions, while unreliable regions are treated more cautiously, for example, by reducing their weights, applying masks, or referencing content from a selected image.

Within the pixel space, these methods can be roughly divided into two categories based on how inconsistencies are handled. Pixel-based approaches compute consistency or confidence at the pixel or patch level and translate them into fusion weights or reliability maps, thereby limiting the influence of inconsistent areas. Registration-based approaches represent another major line within pixel-space deghosting \cite{khan2016simple,shim2020ghosting}. They first align the input images into a common coordinate system using techniques such as feature matching, geometric transformations, or dense motion estimation, and then perform fusion on the aligned results \cite{lee2014ghost}. Overall, these methods rely on consistency modeling and reliability control, and they still serve as strong baselines while providing useful insights for later learning-based approaches.

\subsubsection{Pixel-Based Deghosting Methods}

Pixel-based deghosting methods emphasize reliability control instead of explicit motion estimation. They assess cross-exposure consistency at the pixel or local patch level using cues like intensity or gradient differences, local statistics, or robust cost functions, and then turn these into fusion weights or masks. Regions deemed reliable are used to fully exploit complementary exposure information, while unreliable ones are down-weighted and often filled using a chosen reference image. Many approaches compute reliability across multiple pyramid levels and perform fusion in a coarse-to-fine manner \cite{li2021detail}, with some light detail enhancement and tone adjustment afterward. More recent work mainly improves inconsistency detection, adds stronger structure-aware constraints, and seeks a better balance between efficiency and detail preservation.

RM-HDR \cite{lee2014ghost} follows a robust modeling route. It decomposes the sequence into a low-rank background component and sparse change components, separates motion-related inconsistencies at a global level, and reconstructs the result from the stable component without explicit registration or optical flow. It also uses simple consistency constraints to make the separation more stable.
The structural patch decomposition (SPD) family emphasizes structure consistency during fusion. SPD \cite{ma2017robust} decomposes local patches and prioritizes structure that stays consistent across exposures, which helps to suppress ghosting without alignment but can be slow and may produce halos near strong edges. MSPD \cite{li2020fast} accelerates the pipeline with a multi-scale design and fast approximations and reduces halo artifacts at the cost of some texture smoothing. MESPD \cite{li2021detail} further strengthens edge-preserving structure modeling and introduces adaptive exposure weights to better retain high-frequency details in reliable regions while suppressing unreliable contributions under extreme exposures.

One image in the set $\Omega_f$ is selected as the reference image in \cite{1lizg2014}. All pixels in other images are first classified into consistent pixels and inconsistent pixels. All inconsistent pixels are then corrected using intra-correlation within the non-reference image and inter-correlation between the non-reference and reference images. Unfortunately, it is impossible to always correct all inconsistent pixels using these two types of correlations \cite{1zhengj2013}. An interesting patching algorithm was proposed in  \cite{1zhengj2013} on top of the intra-correlation within the non-reference image and inter-correlation between the non-reference and \mbox{reference images}.   

Overall, pixel-based methods control fusion mainly through reliability estimation without relying on explicit alignment. These methods are simple but their performance needs to be improved.

\subsubsection{Registration-Based Deghosting Methods}

Registration-based deghosting methods attribute most fusion artifacts to geometric misalignment across exposures. They first align non-reference frames to a chosen reference frame and then perform fusion on the aligned stack. A typical pipeline estimates a global or local transform or dense optical flow and warps images or features accordingly \cite{oh2014robust}. To reduce the impact of exposure differences on matching, many methods operate in the gradient, structure, or log-intensity domain. After warping, they often run a residual consistency check and handle occlusions, then down-weight mismatched regions or fall back to reference content to prevent artifacts from spreading. In practice, the goal is not perfect alignment everywhere but alignment that is usable and accompanied by a reliable way to flag failures.

RHDR \cite{oh2014robust} leverages the low-rank background in multi-exposure sequences and uses rank minimization to jointly handle alignment and deghosting, while a sparse term isolates motion and saturated areas for robustness. Li et al. \cite{li2019hybrid} focus on handheld sequences with jitter and local motion, performing coarse alignment with optical flow, detecting unreliable regions via motion variance in superpixels, and refining them with local block matching. Exposure mapping is applied to stabilize matching before final fusion. Registration-based methods thus first establish correspondences and then reduce ghosting by isolating regions where alignment fails. Liu et al. \cite{liu2023unsupervised} propose an unsupervised-learning-based optical flow estimation algorithm in the LDR domain. IMFs \cite{1Grossberg2003,16zhengcb2025} are widely used in \cite{liu2023unsupervised,1lizg2014,1zhengjh2014} to {normalize} differently exposed LDR images.  The algorithm in \cite{liu2023unsupervised} can be adopted to improve the ghost removal algorithms in \cite{1lizg2014,1zhengjh2014}.

\subsection{Ghost Removal in Feature Space}

Deep learning-based deghosting methods formulate multi-exposure fusion or HDR reconstruction as an end-to-end mapping from a bracketed exposure sequence to a ghost-free output. Most approaches adopt encoder--decoder or multi-scale architectures, where each exposure is first encoded into a latent feature space, followed by cross-exposure interaction and final reconstruction.

The key differences among these methods lie in how cross-exposure inconsistencies are handled in the feature space. One line of work explicitly estimates motion or geometric correspondence and performs alignment through warping before fusion. Another line avoids explicit resampling and instead models cross-exposure relations through attention mechanisms or Transformer-style interaction \cite{prabhakar2019fast}. A third line introduces generative modeling to recover unreliable regions caused by occlusion, large motion, or saturation, and refines the output through a reconstruction network.

Based on these design choices, feature-space deghosting methods can be broadly categorized into alignment-based, alignment-free, and generative or hybrid approaches. While most methods are trained in a supervised manner, alternative learning strategies such as self-supervised or consistency-driven learning have also been explored to reduce reliance on ground-truth data \cite{catley2022flexhdr}.

\subsubsection{Alignment-Based Methods}

Alignment-based methods explicitly model cross-exposure correspondence before fusion by estimating motion or geometric relationships and warping images or features to a reference view. By bringing multi-exposure inputs into a shared spatial domain, these methods reduce inconsistencies caused by motion and occlusion, making subsequent fusion and reconstruction more stable \cite{kalantari2017deep}. In this pipeline, alignment is a core component rather than a preprocessing step, and the final output is built upon aligned representations.

A representative starting point is the work by Kalantari et al. \cite{kalantari2017deep}. It uses optical flow to warp short- and long-exposure inputs to a medium-exposure reference; then, a CNN synthesizes HDR from the aligned inputs under supervision. Several follow-up works keep this align-then-reconstruct template but strengthen reconstruction. Yan et al. \cite{yan2019multi} exploit multi-scale representations after alignment so that the network can stabilize coarse structures while recovering fine details, which reduces the impact of small alignment errors. A practical issue is that classical flow can fail under extreme exposure gaps. When flow is inaccurate, fusion may treat warped errors as real content. Prabhakar et al. \cite{prabhakar2019fast} replace traditional optical flow with a stronger learning-based estimator to stabilize alignment, then pass aligned results to the reconstruction network. CRNet \cite{yang2024crnet} still adopts flow-guided warping for explicit alignment and uses an optical flow alignment block based on a pretrained SpyNet to align the input stack before restoration. It then strengthens low-frequency structure and high-frequency details on the aligned representations, improving robustness to residual misalignment and producing sharper textures.

As scenes become harder, the key issue is not only whether alignment can be computed but also which aligned regions should be trusted. FlexHDR \cite{catley2022flexhdr} models alignment and exposure uncertainties so that fusion can down-weight unreliable regions and reduce ghosting. Ren et al. \cite{ren2023robust} improve robustness from the correspondence side. They use a matching volume to search for cross-exposure correspondences while considering both motion and exposure differences, which strengthens the basis for displacement estimation and resampling. Another family of methods treats alignment as learnable sampling. Instead of committing to a single flow field, the network predicts offsets and resamples features directly, which is often more flexible under local motion and structural changes. ADNet \cite{liu2021adnet} follows this idea with feature-level deformable alignment, using a pyramid and cascading strategy to align non-reference features to the reference before fusion. Similarly, Chen \mbox{et al. \cite{chen2023improving}} also employ alignment as a preprocessing step, generating aligned features by learning offsets and feature resampling, then suppressing residual artifacts through enhanced fusion modeling. The principle is unchanged: warped features are produced first, and the fusion backbone operates on aligned representations.

\textls[-15]{Some works reduce alignment failures by redesigning the overall workflow. DDMEF~\cite{tan2023deep}} improves alignability before warping through pre-enhancement, then adds extra training signals to emphasize motion regions during fusion, which stabilizes results under extreme exposure differences. Luo et al. \cite{luo2023multi} embed alignment into attention by learning offsets inside attention and using bidirectional interaction to reduce one-way alignment bias. SAFNet \cite{kong2024safnet} emphasizes reliability in practice. It aligns only recoverable textured regions, while saturated or large-motion regions rely more on fusion and completion, which limits error propagation and also improves efficiency. LRT-HDR \cite{mai2022deep} offers a different perspective after explicit warping. It models reconstruction as low-rank tensor completion and unrolls an iterative solver into a trainable network, which helps to separate stable background from sparse motion- and occlusion-related outliers in the aligned space.

Overall, alignment-based methods have evolved from basic frame alignment to more robust and adaptive correspondence modeling, where both alignment accuracy and reliability are explicitly considered. This enables more stable fusion under large exposure differences and complex motion, reducing artifacts and improving visual consistency.

\subsubsection{Alignment-Free Methods}

Alignment-free methods avoid explicit motion estimation or geometric warping and instead learn to merge multi-exposure information directly in the feature space. In dynamic scenes, misalignment means that the same location may contain different content, and with saturation or occlusion, naive fusion can easily introduce ghosting \cite{qi2025sdf}. To address this, such methods typically avoid any resampling-based warping. Instead, they focus on two key tasks: identifying reliable regions and effectively combining them, while suppressing or completing unreliable areas \cite{yan2022lightweight}.

An early representative is DeepHDR \cite{wu2018deep}, which frames dynamic HDR reconstruction as image translation and learns end-to-end fusion without explicit motion estimation or resampling-based warping. A closely related and influential direction focuses on reliability-aware selection through attention or explicit confidence modeling. AHDRNet~\cite{yan2019attention} systematically introduced reference-guided attention and reduced the contribution of misaligned and saturated regions during fusion, which improves robustness without any motion warping. DAHDRNet \cite{yan2022dual} refines this idea by modeling reliability in both spatial and channel dimensions, so the network not only knows where features are unreliable but also which feature channels should not be amplified. Closely related are hierarchical fusion paradigms. HUNet \cite{yan2022lightweight} adopts a lightweight attention design and an efficient fusion backbone to perform reliability-aware feature selection. HFNet \cite{xiong2021hierarchical} progressively integrates multi exposure information through stage-wise fusion with regional confidence modeling. GSANet \cite{li2022gamma} groups the input LDR images with their gamma-mapped counterparts and applies spatial attention to select reliable regions. Despite different implementations, they share the same principle of suppressing inconsistent content first and then performing fusion and reconstruction.

When local attention is not enough for large displacements, later methods bring in stronger global reasoning or long-term memory. HyHDRNet \cite{yan2023unified} adopts an alignment-free, patch-level aggregation strategy with ghost-aware attention and gating, enabling reliable cross-exposure interaction without explicit warping and improving structural stability under large motion. NHDRRNet \cite{yan2020deep} uses non-local correlation to perform global evidence selection in deep feature space, allowing each pixel to draw from the most trustworthy exposure cues under large motion and occlusion. HDRRNN \cite{prabhakar2021self} takes a different angle by reusing information over time with a lightweight self-gated memory recurrent unit, accumulating stable cross-exposure cues while keeping the parameter budget small. GCANet \cite{yan2022high} uses gradient guidance and context aggregation, treating structural edges as a more stable cue for fusion and improving detail fidelity without registration. These models can be viewed as representative alignment-free CNN solutions, sharing the same objective as later Transformer-based methods while relying on convolution and context aggregation to realize it.

The introduction of Transformers further strengthens long-range dependency modeling, while the core remains implicit alignment and reliability-aware fusion. HDR-Transformer \cite{liu2022ghost} adopts a dual-branch design to capture global and local dependencies, using global context to stabilize structure under large motion and local modeling to protect textures. Selective TransHDR \cite{song2022selective} focuses computation where it matters by first identifying ghost-prone regions and then applying Transformer reasoning selectively, improving detail and color while controlling cost. SCTNet \cite{tel2023alignment} feeds multi-exposure features directly into a Transformer and adds semantic consistency constraints to stabilize cross-exposure relationships, making the model more confident in suppressing misaligned content under large motion. HL-HDR \cite{zhang2024hl} splits the problem by separating low-frequency structure and high-frequency details in feature space. Lightweight convolution emphasizes local textures while the Transformer handles global structure, reducing computation without sacrificing deghosting quality. RFG-HDR \cite{lee2024rfg} introduces contrastive learning to disentangle exposure-related global cues from exposure-invariant structural cues, and uses these representative features to guide Transformer fusion so that cross-exposure interactions become more stable and controllable. To address the overhead of Transformers, EiffHDR~\cite{zhang2024eiffhdr} replaces heavy attention with lightweight gating and large-kernel convolutions while retaining implicit alignment and multi-scale modeling, reflecting a broader shift toward balancing efficiency and performance.

Some methods avoid misalignment contamination by redesigning the workflow rather than only changing the fusion block. Chung et al. \cite{chung2022high} weaken cross-exposure inconsistency through adaptive exposure or luminance adjustment and then complete saturated regions, recovering missing details without explicit motion compensation. APNT-Fusion~\cite{chen2022attention} uses a dual-stream framework where one stream suppresses inconsistent content with motion and saturation attention and the other transfers texture into saturated regions via multi-scale feature matching, followed by progressive mixing to jointly improve deghosting and detail recovery. SAMHDR \cite{li2024single} first reconstructs a relatively clean single-image HDR from the reference exposure as a structural anchor and then uses it to constrain multi-exposure fusion so that the network borrows detail from other exposures without inheriting incorrect motion. AFUNet \cite{li2025afunet} formulates alignment and fusion as a trainable alternating optimization process, using reference features to guide cross-exposure interaction at each iteration and progressively correcting inconsistencies during fusion rather than relying on pre-alignment. iHDR \cite{yuan2025ihdr} decomposes fusion into pairwise merging with iterative accumulation, using structure cues and difference masks to maintain consistency and attention-based implicit fusion to naturally support an arbitrary number of inputs while continuously suppressing ghosting.  DeepDuoHDR \cite{alpay2024deepduohdr} adopts a two-exposure, patch-level deghosting strategy for mobile HDR imaging, predicting aligned low-exposure patches only for saturated regions and merging them with the high-exposure reference to reduce ghosting with lower computational cost.

These methods share a common idea. Rather than warping inputs with optical flow or geometric registration, they select and combine reliable evidence directly in feature space. Early approaches use reference-guided attention or confidence modeling to handle misalignment and saturation. Later methods add long-range reasoning with non-local correlation, memory mechanisms, and Transformers to stabilize structures under large motion. Recent work further tackles difficult cases by separating structure and detail, stabilizing cross-exposure relations with contrastive or semantic constraints, and gradually correcting inconsistencies through iterative fusion. Overall, they move deghosting from geometric alignment to robust evidence selection, enabling cleaner HDR reconstruction under motion, occlusion, and saturation. Similarly to the ghost removal algorithm in pixel space \cite{1lizg2014}, {these algorithms could be improved by adding one step to correct \mbox{inconsistent features}.}

\subsubsection{Generative and Other Methods}

\textls[-5]{Generative and hybrid methods view HDR deghosting as a content recovery task rather than simple regression-based fusion. In dynamic scenes, motion, occlusion, and saturation often lead to missing or corrupted regions, where direct fusion can introduce visible artifacts. To address this,} generative models provide stronger priors to fill in these unreliable areas with plausible content. Most methods follow a similar strategy: reliable regions are fused directly while unreliable ones are handled by a generative module, followed by refinement to ensure consistent geometry, exposure, and color across \mbox{the image}.

HDR-GAN \cite{niu2021hdr} treats HDR reconstruction as conditional generation using adversarial training to complete occluded and saturated regions. It does not rely on pre-alignment but uses reference-guided feature interaction and multi-scale supervision with fully supervised paired data. DiffHDR \cite{yan2023toward} applies diffusion models, conditioning denoising on implicitly aligned features and using feature modulation to reduce ghosting and color drift, with noise estimation and image-space constraints to improve fidelity in saturated areas.

Because diffusion sampling can be expensive, later work often adopts hybrid designs that use diffusion only where it matters most. UltraFusion \cite{chen2025ultrafusion} follows the same idea. It first produces a stable base result through aligned multi exposure fusion, then performs conditional diffusion completion in highlight missing or saturated regions under an unreliability mask, avoiding a direct blend of conflicting evidence. However, since UltraFusion mainly focuses on extreme exposure differences in static scenes, its robustness under large real motion may still be limited. Zhang et al. \cite{zhang2024efficient} embed diffusion into a regression backbone so that diffusion provides structural priors in challenging regions while the regression network efficiently reconstructs the rest, balancing speed and detail. LFDiff \cite{hu2024generating} pushes this idea further by restricting diffusion to a low-frequency structure, then using a regression network to restore high-frequency details, reducing runtime while preserving the benefit of generative priors in hard regions. Yan et al. \cite{yan2024dynamic} locate motion-occluded and overexposed regions via semantic segmentation, perform stepwise diffusion completion, and refine fusion, effectively handling extreme motion and saturation where alignment fails.

Overall, this family evolves from GAN-driven completion to diffusion-driven generation, and then toward efficient hybrid pipelines that reserve generative modeling for the hardest regions under motion and saturation.

In addition to generative modeling, some methods explore alternative learning paradigms such as unsupervised or self-supervised learning, where supervision is derived from consistency or reconstruction rather than paired ground truth.

When reliable HDR ground truth is not available, deghosting in dynamic scenes cannot rely on direct regression. Instead, many methods shift supervision from the output to the imaging process, requiring the predicted HDR to both look reasonable and explain the input exposures \cite{yan2023smae}. A common strategy is to generate an HDR result, map it back to each exposure using differentiable camera or exposure models, and train with reconstruction consistency. Uncertain regions caused by motion, occlusion, or saturation are usually handled with masks or confidence maps to avoid mixing conflicting evidence.

Following this idea, UPHDR-GAN \cite{li2022uphdr} removes the need for paired data by combining adversarial learning with local consistency constraints. FSHDR \cite{prabhakar2021labeled} further leverages unlabeled sequences, using a small labeled set for guidance and building self-supervised signals through re-rendering, achieving strong performance even with limited data. To improve stability with scarce labels, SMAE \cite{yan2023smae} adopts a two-stage design: it first restores saturated regions via self-supervision, then performs semi-supervised deghosting using only reliable pseudo-labels. SelfHDR \cite{zhang2023self} decomposes the learning target into more controllable components, using color cues and a reference exposure to guide structure and reduce artifacts. In contrast, DSFMEF \cite{zhao2025single} tackles the problem at the data level by generating a consistent multi-exposure sequence from a single RAW image, reducing misalignment before fusion.

\subsection{Critical Comparison and Trade-Offs}
The taxonomy discussed above suggests that different HDR reconstruction methods do not simply differ in implementation details; they rely on different assumptions about exposure, motion, and reliable image content. Pixel-space methods are relatively transparent because they work directly with image intensities, local structures, or fusion weights. This makes them efficient and easy to understand in conventional MEF pipelines. However, the reliability measures used in these methods are often designed manually or estimated from local observations. When the exposure gap is large, or when occlusion and complex motion destroy local correspondence, such pixel-level cues may become unreliable. Feature-space methods address this problem from another direction. They learn representations for exposure correction, alignment, and fusion, which gives them better flexibility in recovering missing details and suppressing ghosting. The price is that their behavior depends more on training data, network architecture, and supervision. In some difficult cases, they may also produce inconsistent luminance or plausible-looking textures that are not \mbox{physically accurate}.

In dynamic HDR reconstruction, the central question is whether cross-exposure correspondence should be explicitly built before fusion. Optical-flow-based alignment provides a clear solution and is effective when motion is not too large and exposure differences are moderate. Once the corresponding regions are saturated, strongly underexposed, or occluded, the estimated flow may no longer be trustworthy. Deformable alignment offers more flexibility by learning local offsets in the feature space, which is useful for non-rigid motion and residual misalignment. Still, inaccurate offsets can also transfer inconsistent structures into the reconstructed result. Alignment-free methods avoid this explicit warping step. Instead, they rely on attention, confidence estimation, non-local matching, recurrent memory, or Transformer-based interaction to select useful information across exposures. These methods are less tied to optical flow, but they are not free from errors: their performance depends on whether the model can correctly identify unreliable regions and suppress conflicting content. Therefore, alignment-based methods are more suitable when correspondences can be estimated reliably, while alignment-free methods are generally more attractive in scenes with severe saturation, occlusion, or large \mbox{exposure difference}s.

The same trade-off also appears at the architecture and supervision levels. CNN-based methods remain practical because they are efficient and good at local texture recovery, but they may have difficulty maintaining global luminance consistency in scenes with large exposure variation. Transformer-based methods model long-range dependencies more naturally and can capture global exposure relationships, although this usually leads to higher memory and computational cost. Deterministic regression methods tend to be stable and easier to evaluate, whereas GAN- or diffusion-based methods are better suited to completing heavily saturated or occluded regions, but may generate details that are visually convincing rather than physically faithful. Supervised methods benefit from paired HDR references, yet such references are hard to obtain for dynamic scenes. Self-supervised, unsupervised, and no-reference methods are easier to extend to real-world data, but their objectives may not fully correspond to true HDR reconstruction quality. From a practical viewpoint, high-quality reconstruction and deployment efficiency are still difficult to achieve at the same time. Large Transformer or diffusion models are useful for challenging scenes, whereas lightweight CNNs, lookup-table-based methods, selective alignment, and region-aware processing are more suitable for mobile photography and embedded vision.

\section{Evaluation of Multi-Shot HDR Imaging Algorithms}
\subsection{Datasets}

Multi-exposure image sequences are the basis of MEF research, providing both benchmarks for comparing methods and data for training and evaluating deep learning models. Existing datasets vary widely in exposure levels, scale, resolution, imaging quality, and scene diversity. Some generate LDR inputs using simulated camera response functions or CRF databases, while others offer real LDR sequences with corresponding HDR references. To help readers, this paper summarizes representative public datasets. Table~\ref{tab:datasets} lists dataset names, publication venues, data types, splits, and sequence or image counts.

\begin{table}[H]
\small
\centering
\caption{Datasets for comparison of MEF algorithms. Note: T, Te, and val denote training set, test set, and validation set, respectively.} 
\label{tab:datasets}

\begin{adjustwidth}{-\extralength}{0cm}
\begin{tabularx}{\fulllength}{cCCccC}
\toprule
\textbf{Name} & \textbf{Source} & \textbf{Type} & \textbf{Split} & \textbf{Description} & \textbf{Data Source} \\ 
Tel \cite{tel2023alignment} & ICCV-2023 & Dynamic & 108T + 36Te & 144 sequences, 432 images & Real \\ 
MobileHDR \cite{liu2023joint} & CVPR-2023 & Dynamic & 223T + 28Te & 251 sequences & Real \\ 
Canon5D4 \cite{xu2021deep} & TCSVT-2021 & Static &{300T + 100Te + 100val} & 500 sequences & Real \\
NTIRE\_2021 \cite{perez2021ntire} & CVPRW-2021 & Dynamic & 1494T + 201Te + 60Val & {1755 sequences} & Synthetic \\ 
IISc$_V$AL \cite{prabhakar2021self} & TCI-2021 & Dynamic & 70T + 14Te & 84 sequences, 588 images & Real \\ 
MEFB \cite{zhang2021benchmarking} & IF-2021 & Static & 100Te & 100 sequences, 200 images & Real \\
DeghostingIQA \cite{fang2019image} & TIP-2019 & Dynamic & 20Te & 20 sequences, 180 images & Real \\
SICE \cite{cai2018learning} & TIP-2018 & Static & 412T + 59Te + 118val & 589 sequences, 4413 images & Real \\
MEF-IQA \cite{ma2017multi} & TCI-2018 & Static & 192Te & 24 sequences, 192 images & Real \\ 
Kalantari \cite{kalantari2017deep} & TOG-2017 & Dynamic & 74T + 15Te & 89 sequences, 267 images & Real \\
DeepFuse Dataset \cite{ram2017deepfuse} & ICCV-2017 & Static & {75T + 25Te} & 100 sequences  & Real \\
Hu \cite{hu2013hdr} & CVPR-2013 & Dynamic & 85T + 15Te & 100 sequences, 300 images & Synthetic \\ 
Sen \cite{sen2012robust} & TOG-2012 & Dynamic & 8Te & 8 sequences & Real \\ \bottomrule
\end{tabularx}%
\end{adjustwidth}
\end{table}

These datasets can be broadly grouped into static and dynamic types. Static datasets, such as SICE, MEFB, MEF-IQA, DeepFuse Dataset, and Canon5D4, are more suitable for static MEF evaluation, detail enhancement, luminance consistency, and tone-mapped quality assessment. Dynamic datasets, such as Kalantari, Hu, Sen, Tel, IISc$_V$AL, DeghostingIQA, and MobileHDR, are more suitable for motion compensation, saturation recovery, dynamic HDR reconstruction, and deghosting evaluation. However, existing datasets still have clear limitations. Static datasets cannot evaluate motion-induced ghosting, while many dynamic benchmarks contain limited motion patterns, moderate exposure gaps, or relatively simple occlusions. Large-scale synthetic datasets such as NTIRE\_2021 are useful for supervised training, but may suffer from a domain gap with real camera noise, ISP processing, and handheld capture. Therefore, current benchmarks still lack sufficient sequences that simultaneously contain large exposure ratios, severe over- or underexposure, complex object motion, and cross-exposure occlusion, which partly explains why high benchmark scores do not always translate into robust real-world HDR imaging performance.

\subsection{Evaluation Metrics}
The quality assessment of multi-exposure fusion results is generally divided into subjective and objective evaluations \cite{jia2017blind}. Subjective evaluation reflects human perception, including exposure naturalness, detail clarity, color fidelity, and artifact presence. However, it is influenced by evaluator experience, display conditions, and task preferences, making reproducibility difficult.

\subsubsection{PSNR}
Peak signal-to-noise ratio (PSNR) \cite{endo2017deep,wu2018deep} is a classic full-reference metric for measuring pixel-level error between reconstructed results and reference images, fundamentally based on the mean squared error (MSE). Given the predicted HDR image $\hat{\mathbf{H}} \in \mathbb{R}^{M\times N\times C}$ and the ground truth $\mathbf{H}$, the MSE and PSNR are defined as
\begin{equation}
\mathrm{MSE}(\hat{\mathbf{H}},\mathbf{H})
=\frac{1}{MNC}\left\lVert \hat{\mathbf{H}}-\mathbf{H}\right\rVert_{F}^{2},
\end{equation}
\begin{equation}
\mathrm{PSNR}(\hat{\mathbf{H}}, \mathbf{H})
=10\log_{10}\left(\frac{\mathcal{R}^{2}}{\mathrm{MSE}(\hat{\mathbf{H}}, \mathbf{H})}\right),
\end{equation}
where $\mathcal{R}$ denotes the peak dynamic range. In practice, $\mathcal{R}=1$ when images are normalized to $[0,1]$.

Higher PSNR values usually indicate smaller reconstruction errors. However, PSNR only measures pixel-wise differences and correlates weakly with perceived structure, luminance, and color distortions in MEF. Therefore, raw PSNR is rarely reported directly; instead, PSNR-L in the linear domain and PSNR-$\mu$ in the $\mu$-law tone-mapped domain are commonly adopted.

For datasets providing HDR ground truth in linear radiance, PSNR-L is computed directly in the linear domain. To account for HDR's wide dynamic range and the approximately nonlinear luminance sensitivity of human vision, PSNR-$\mu$ applies $\mu$-law tone mapping to both $\hat{\mathbf{H}}$ and $\mathbf{H}$ before computing PSNR. The $\mu$-law mapping is defined as
\begin{equation}
\mathcal{T}_{\mu}(x)=\frac{\log(1+\mu x)}{\log(1+\mu)}, \quad \mu = 5000.
\end{equation}
Accordingly, PSNR-L and PSNR-$\mu$ are formulated as
\begin{align}
\mathrm{PSNR\mbox{-}L} &= \mathrm{PSNR}(\hat{\mathbf{H}}, \mathbf{H}), \\
\mathrm{PSNR\mbox{-}\mu} &= \mathrm{PSNR}\!\left(\mathcal{T}_{\mu}(\hat{\mathbf{H}}),\,\mathcal{T}_{\mu}(\mathbf{H})\right).
\end{align}

\subsubsection{SSIM}
The structural similarity index measure (SSIM) \cite{wang2004image,hanhart2015benchmarking} is a full-reference perceptual metric that evaluates the structural consistency between a prediction and its reference. Unlike PSNR, which is driven by pixel-wise errors, the SSIM measures similarity within local windows by jointly considering luminance, contrast, and structural components. Given the predicted HDR image $\hat{\mathbf{H}}$ and the ground truth $\mathbf{H}$, the SSIM is defined as
\begin{equation}
\mathrm{SSIM}(\hat{\mathbf{H}}, \mathbf{H})
= \frac{(2\mu_{\hat{H}}\mu_{H}+C_{1})(2\sigma_{\hat{H}H}+C_{2})}
{(\mu_{\hat{H}}^{2}+\mu_{H}^{2}+C_{1})(\sigma_{\hat{H}}^{2}+\sigma_{H}^{2}+C_{2})},
\end{equation}
where $\mu_{\hat{H}}$, $\mu_{H}$, $\sigma_{\hat{H}}^{2}$, $\sigma_{H}^{2}$, and $\sigma_{\hat{H}H}$ are the local mean, variance, and covariance computed within a window. The constants $C_1=(K_1\mathcal{R})^2$ and $C_2=(K_2\mathcal{R})^2$ stabilize the computation, typically with $K_1=0.01$ and $K_2=0.03$, where $\mathcal{R}$ denotes the peak dynamic range and is set to 1 for images normalized to $[0,1]$. A higher SSIM indicates better structural similarity.

The SSIM is typically reported as SSIM-L in the linear radiance domain and as SSIM-$\mu$ after applying $\mu$-law tone mapping. For datasets with HDR ground truth in linear radiance, SSIM-L is computed directly between $\hat{\mathbf{H}}$ and $\mathbf{H}$. To better align with perceptually relevant comparisons under HDR's wide dynamic range, SSIM-$\mu$ applies the same $\mu$-law mapping to both the prediction and the ground truth before computing the SSIM. Their formulas are as~follows:\vspace{+6pt}
\begin{align}
\mathrm{SSIM\mbox{-}L} &= \mathrm{SSIM}(\hat{\mathbf{H}}, \mathbf{H}), \\
\mathrm{SSIM\mbox{-}\mu} &= \mathrm{SSIM}\!\left(\mathcal{T}_{\mu}(\hat{\mathbf{H}}),\,\mathcal{T}_{\mu}(\mathbf{H})\right).
\end{align}

\subsubsection{HDR-VDP-2}
HDR-VDP-2 \cite{mantiuk2011hdr} is a perceptually motivated full-reference metric for HDR image quality assessment. It compares the reconstructed HDR image with the ground truth by simulating key stages of the human visual system and predicting the visibility of differences under HDR viewing conditions. In general, a higher HDR-VDP-2 score indicates better perceptual quality and closer agreement with the reference, making it a commonly adopted metric for datasets that provide HDR ground truth.

\subsubsection{MEF-SSIM}
Multi-exposure fusion structural similarity (MEF-SSIM) \cite{ma2015perceptual} is a source-reference/no-reference metric for multi-exposure fusion, since it evaluates the fused image using the input exposure sequence rather than an HDR ground-truth reference. Let \(F\) denote the fused image and \(\{S_i\}_{i=1}^{K}\) denote the input exposure sequence. MEF-SSIM measures how well \(F\) preserves local structural details from \(\{S_i\}_{i=1}^{K}\). The image is divided into local blocks. For the \(j\)-th block, a desired structure block \(\hat{\mathbf{s}}^{(j)}\) is estimated from the input exposure sequence, and the structural similarity is computed between \(\hat{\mathbf{s}}^{(j)}\) and the corresponding fused block \(\mathbf{f}^{(j)}\). A larger value implies better structure preservation. It is defined as
\begin{equation}
S(\hat{\mathbf{s}},\mathbf{f})
=
\frac{2\sigma_{\hat{\mathbf{s}}\mathbf{f}}+C}
{\sigma_{\hat{\mathbf{s}}}^{2}+\sigma_{\mathbf{f}}^{2}+C},
\end{equation}
\begin{equation}
Q(F)
=
\frac{1}{M}\sum_{j=1}^{M}
S\!\left(\hat{\mathbf{s}}^{(j)},\mathbf{f}^{(j)}\right),
\end{equation}
where \(F\) denotes the fused image and \(\{S_i\}_{i=1}^{K}\) denotes the input exposure sequence. The vector \(\mathbf{f}^{(j)}\) is the \(j\)-th local block of \(F\), while \(\hat{\mathbf{s}}^{(j)}\) denotes the desired structure block estimated from the input exposure sequence. \(M\) is the number of local blocks. \(\sigma_{\hat{\mathbf{s}}}^{2}\) and \(\sigma_{\mathbf{f}}^{2}\) are the local variances of \(\hat{\mathbf{s}}^{(j)}\) and \(\mathbf{f}^{(j)}\), respectively; \(\sigma_{\hat{\mathbf{s}}\mathbf{f}}\) is their local covariance; and \(C\) is a small constant for numerical stability.

\subsubsection{MI}
Mutual information (MI) \cite{qu2002information} measures the amount of shared information between the fused image and the source images, reflecting how much information from the inputs is preserved in the fusion result. Let \(F\) denote the fused image and \(\{S_i\}_{i=1}^{K}\) denote the input exposure sequence. A larger MI value usually indicates richer retained source information. It is defined as
\begin{equation}
\mathrm{MI}(F,S_i)
=
\sum_{f,s}
p_{F,S_i}(f,s)
\log_{2}
\frac{p_{F,S_i}(f,s)}
{p_F(f)p_{S_i}(s)},
\end{equation}
\begin{equation}
\mathrm{MI}(F,\{S_i\}_{i=1}^{K})
=
\sum_{i=1}^{K}\mathrm{MI}(F,S_i),
\end{equation}
where \(p_{F,S_i}(f,s)\) denotes the joint probability distribution between the fused image \(F\) and the \(i\)-th source image \(S_i\), and \(p_F(f)\) and \(p_{S_i}(s)\) are the corresponding marginal probability distributions. The accumulated mutual information \(\mathrm{MI}(F,\{S_i\}_{i=1}^{K})\) measures the amount of source information transferred from the input exposure sequence to the fused image.

\subsubsection{SD}
Standard deviation (SD) \cite{haghighat2011non} describes the spread of gray levels in the fused image and is often used to reflect image contrast. A larger SD value usually indicates higher contrast. It is computed as\vspace{+6pt}
\begin{equation}
\sigma=
\sqrt{\sum_{i=0}^{L-1}\left(i-\bar{i}\right)^{2}h_{F}(i)},
\qquad
\bar{i}=\sum_{i=0}^{L-1} i\,h_{F}(i),
\end{equation}
where $L$ is the number of gray levels, $i \in \{0,\ldots,L-1\}$ denotes the gray-level index, and $h_F(i)$ is the normalized histogram of the fused image at gray level $i$.

\subsubsection{Entropy (EN)}
Entropy (EN), also denoted as $H_e$ \cite{naidu2010discrete}, measures the information content of the fused image based on its gray-level distribution. A larger EN value usually indicates richer information. It is defined as
\begin{equation}
\mathrm{EN}
=
-\sum_{i=0}^{L-1} h_{F}(i)\log_{2}h_{F}(i),
\end{equation}
where $h_F(i)$ denotes the normalized histogram of the fused image.

\subsubsection{$Q^{AB/F}$}
$Q^{AB/F}$ \cite{xydeas2000objective} is a no-reference fusion metric that evaluates how well the fused image preserves edge information from the source images. A larger value usually indicates better edge preservation and thus better fusion performance. It is computed as
\begin{equation}
Q^{AB/F}
=
\frac{
\sum\limits_{i=1}^{N}\sum\limits_{x=1}^{H}\sum\limits_{y=1}^{W}
W^{i}(x,y)\,Q^{i}_{g}(x,y)\,Q^{i}_{\alpha}(x,y)
}{
\sum\limits_{i=1}^{N}\sum\limits_{x=1}^{H}\sum\limits_{y=1}^{W}
W^{i}(x,y)
},
\end{equation}
where $N$ is the number of source images and $H\times W$ is the image size. $Q^{i}_{g}(x,y)$ denotes the edge strength preservation at pixel $(x,y)$ from the $i$-th source image, $Q^{i}_{\alpha}(x,y)$ denotes the edge orientation preservation, and $W^{i}(x,y)$ is a weighting term related to gradient magnitude. In general, a higher $Q^{AB/F}$ indicates that more edge details from the sources are retained in the fused result.

\subsubsection{NIQE}
Natural image quality evaluator (NIQE) \cite{mittal2012making} is a no-reference metric based on natural scene statistics. It measures how much an image deviates from high-quality natural images and does not require a reference or training on distorted examples. Lower NIQE scores indicate more natural-looking results and higher perceptual quality.

\subsubsection{Additional Metrics Used in Quantitative Comparisons}
In addition to the commonly used metrics introduced above, several auxiliary metrics are also adopted in quantitative comparisons of multi-exposure fusion and HDR imaging methods. These metrics evaluate global correlation, sharpness, information transfer, feature preservation, perceptual fidelity, or fusion distortion from different perspectives. Some of them complement full-reference evaluation while others are useful when HDR ground truth is unavailable. Therefore, they should be interpreted together with task requirements and visual inspection rather than used as independent evidence of overall quality.

Correlation coefficient (CC) measures the global linear correlation between the fused image and the source or reference image. It is useful for checking whether the fused result preserves the overall intensity distribution of the inputs, but it is insensitive to local artifacts, ghosting, and perceptual distortions. Average gradient (AG) reflects image sharpness by measuring local intensity changes. A higher AG often indicates richer edges and textures, but it may also be increased by noise, halos, ringing, or over-sharpening. Normalized mutual information (NMI) and feature mutual information (FMI) are extensions of mutual information. NMI reduces the influence of different entropy levels between images while FMI evaluates information transfer in feature domains such as gradient, edge, wavelet, or phase-congruency representations. These metrics are useful for measuring information preservation but they do not distinguish useful structures from artifacts or noise.

Visual information fidelity (VIF) measures how much visual information is preserved with respect to a reference image based on natural scene statistics. It is more perceptually meaningful than purely pixel-wise errors, but it still requires a reliable reference and is not specifically designed for HDR deghosting. Multi-scale structural similarity (MS-SSIM) extends the SSIM to multiple resolutions and is useful for evaluating scale-dependent structural consistency. However, it may still be insensitive to color shifts, exposure naturalness, and local ghosting artifacts. Nonlinear correlation information entropy (QNCIE) evaluates nonlinear correlation and information preservation among source and fused images, but its interpretation is less direct than PSNR, SSIM, or MEF-SSIM. The Chen--Varshney metric ($Q_{CV}$) measures local visual information loss between source images and the fused result, where lower values usually indicate better fusion quality. However, it may penalize perceptually acceptable contrast changes.

The tone-mapped image quality index (TMQI) is mainly used to evaluate display-ready tone-mapped HDR images by combining structural fidelity and naturalness. It is suitable when the final output is intended for visual display, but it does not directly measure radiance-domain accuracy or deghosting performance in dynamic multi-exposure scenes. Overall, these auxiliary metrics provide complementary evidence from different perspectives, but none of them can fully characterize HDR reconstruction quality, exposure correctness, color fidelity, and ghost removal at the same time.

\subsubsection{Discussion on Metric Suitability and Limitations}
The selection of evaluation metrics should depend on the output form and the target task. For HDR reconstruction with ground-truth radiance maps, full-reference metrics such as PSNR-L, PSNR-$\mu$, SSIM-L, SSIM-$\mu$, MS-SSIM, HDR-VDP-2, VIF, and TMQI are commonly used to evaluate pixel fidelity, structural consistency, perceptual distortion, or tone-mapped visual quality. However, these metrics rely on reliable references and may be affected by spatial misalignment, tone-mapping functions, or display assumptions.

For multi-exposure fusion without HDR ground truth, source-reference or no-reference metrics such as MEF-SSIM, CC, AG, EN, MI, NMI, FMI, QNCIE, $Q^{AB/F}$, and $Q_{CV}$ are more frequently adopted. These metrics evaluate structural preservation, information transfer, sharpness, nonlinear correlation, or fusion distortion from different perspectives. Nevertheless, high scores do not always indicate visually pleasing HDR results, since noise, halos, over-sharpening, color shifts, or unnatural contrast may also improve some \mbox{statistical indicators}.

In dynamic scenes, deghosting evaluation is more challenging. Ghosting artifacts are usually local and visually salient, but their contribution to global quantitative scores may be limited. Therefore, a method may obtain competitive PSNR, SSIM, MEF-SSIM, or information-theoretic scores while still producing visible double edges, motion residues, or local structural duplications. As summarized in Tables~\ref{tab:hdr_perceptual_metrics} and~\ref{tab:mef_fusion_metrics}, no single metric can comprehensively evaluate multi-exposure HDR imaging. Quantitative results should therefore be interpreted together with visual comparison, local inspection of motion, saturation, and occlusion regions, and perceptual assessment whenever possible.

\begin{table}[H]\renewcommand\arraystretch{1.3}
\fontsize{8}{8}\selectfont
\caption{Summary of HDR reconstruction and perceptual quality metrics. Note: FR and NR denote full-reference and no-reference metrics, respectively. GT denotes ground truth. ($\uparrow$) means higher is better, while ($\downarrow$) means lower is better.}

\label{tab:hdr_perceptual_metrics}

\begin{adjustwidth}{-\extralength}{0cm}
\begin{tabularx}{\linewidth}{
>{\centering\arraybackslash}p{0.1\linewidth}
>{\centering\arraybackslash}p{0.10\linewidth}
>{\centering\arraybackslash}p{0.18\linewidth}
>{\raggedright\arraybackslash}p{0.22\linewidth}
>{\raggedright\arraybackslash}p{0.27\linewidth}
}
\toprule
\textbf{Metric} & \textbf{Type / Dir.} & \textbf{Input} & \multicolumn{1}{c}{\textbf{Suitable Tasks}} & \multicolumn{1}{c}{\textbf{Main Weakness}} \\
\midrule
PSNR/PSNR-L & FR/($\uparrow$) & Linear HDR & Radiance-domain HDR reconstruction with GT & Weak perceptual correlation; insensitive to local ghosting. \\\midrule
PSNR-\(\mu\) & FR/($\uparrow$) & Tone-mapped HDR & Wide-range HDR evaluation after perceptual compression & Depends on tone mapping; may hide radiance-domain errors. \\\midrule
SSIM/SSIM-L & FR/($\uparrow$) & Linear HDR or LDR & Structural fidelity assessment with reference & Weak for color shift and localized ghosting artifacts. \\\midrule
SSIM-\(\mu\) & FR/($\uparrow$) & Tone-mapped HDR & Perceptual structural comparison & Still limited for HDR perception and ghost artifacts. \\\midrule
MS-SSIM & FR/($\uparrow$) & HDR or tone-mapped HDR & Multi-scale structural consistency & May overlook exposure naturalness and color fidelity. \\\midrule
HDR-VDP-2 & FR/($\uparrow$) & HDR & HDR perceptual quality assessment & Requires HDR reference and viewing/display assumptions. \\\midrule
TMQI & FR/($\uparrow$) & Tone-mapped HDR & Display-ready tone-mapped image evaluation & Mainly designed for tone mapping; not deghosting-specific. \\\midrule
VIF & FR/($\uparrow$) & Reference and test images & Visual information fidelity assessment & Measures information fidelity but is not ghosting-specific. \\\midrule
NIQE & NR/($\downarrow$) & Fused LDR or tone-mapped image & Blind naturalness assessment & Based on natural image statistics; not HDR/MEF-specific. \\
\bottomrule
\end{tabularx}
\end{adjustwidth}

\end{table}

\vspace{-12pt}
\begin{table}[H]\renewcommand\arraystretch{1.3}
\fontsize{8}{8}\selectfont
\caption{Summary of source-reference and no-reference metrics for multi-exposure fusion. Note: FR, SR, and NR denote full-reference, source-reference, and no-reference metrics, respectively. ($\uparrow$) means higher is better, while ($\downarrow$) means lower is better.}
\label{tab:mef_fusion_metrics}

\begin{adjustwidth}{-\extralength}{0cm}
\begin{tabularx}{\fulllength}{
>{\raggedright\arraybackslash}p{0.09\linewidth}
>{\centering\arraybackslash}p{0.10\linewidth}
>{\raggedright\arraybackslash}p{0.19\linewidth}
>{\raggedright\arraybackslash}p{0.22\linewidth}
>{\raggedright\arraybackslash}p{0.28\linewidth}
}
\toprule
\textbf{Metric} & \textbf{Type / Dir.} & \textbf{Input} & \textbf{Suitable Tasks} & \textbf{Main Weakness} \\
\midrule
MEF-SSIM & SR/NR/($\uparrow$) & Input exposure sequence and fused LDR & Static multi-exposure fusion & Weak for ghosting, color fidelity, and radiance accuracy. \\\midrule
CC & FR/SR/($\uparrow$) & Source/reference image and fused image & Global similarity or source preservation & Ignores local artifacts and perceptual distortions. \\\midrule
AG & NR/($\uparrow$) & Fused LDR & Sharpness and detail evaluation & May favor noise, halos, ringing, or over-sharpening. \\\midrule
EN & NR/($\uparrow$) & Fused LDR & Information richness estimation & High entropy does not ensure natural or artifact-free results. \\\midrule
MI & SR/NR/($\uparrow$) & Source images and fused image & Information transfer evaluation & Cannot separate useful details from noise or artifacts. \\\midrule
NMI & SR/NR/($\uparrow$) & Source images and fused image & Normalized information transfer evaluation & Weak for perceptual quality and ghosting assessment. \\\midrule
FMI & SR/NR/($\uparrow$) & Source/fused feature maps & Feature-level information preservation & Depends on the selected feature representation. \\\midrule
QNCIE & SR/NR/($\uparrow$) & Source images and fused image & Nonlinear correlation and entropy evaluation & Less interpretable and relatively more computationally complex. \\\midrule
(Q\textsuperscript{{AB/F}}) & SR/NR/($\uparrow$) & Source images and fused image & Edge preservation in image fusion & Edge preservation does not ensure correct exposure or color. \\\midrule
CE & SR/NR/($\downarrow$) & Source images and fused image & Distribution difference or information loss estimation & Histogram-based; ignores spatial structure. \\\midrule
(Q\_P) & SR/NR/($\uparrow$) & Source images and fused image & Perceptual fusion quality assessment & Not specific to HDR radiance fidelity or dynamic ghosting. \\\midrule
(Q\_W) & SR/NR/($\uparrow$) & Source images and fused image & Structural transfer assessment & Less sensitive to exposure correctness. \\\midrule
(Q\_{CB}) & SR/NR/($\uparrow$) & Source images and fused image & HVS-inspired visual fusion evaluation & Not designed for HDR reconstruction or dynamic deghosting. \\\midrule
(Q\_{CV}) & SR/NR/($\downarrow$) & Source images and fused image & Fusion distortion or visual loss estimation & May penalize perceptually acceptable contrast changes. \\
\bottomrule
\end{tabularx}
\end{adjustwidth}

\end{table}


\subsection{Quantitative Comparison}
To evaluate different MEF algorithms, quantitative comparisons are conducted on the SICE and MEFB datasets using PSNR, SSIM, CC, AG, VIF, MEF-SSIM, TMQI, and MS-SSIM.

Table~\ref{tab:sice} shows that deep learning-based fusion methods exhibit clear performance differences on the SICE dataset. Early CNN-based models achieve moderate results, while more recent methods provide consistent improvements. These gains indicate enhanced capability in preserving structural details and maintaining global luminance consistency.

\begin{table}[H]
\caption{Quantitative comparison of MEF algorithms on the SICE dataset. $\uparrow$ means higher is better. The best results are highlighted in bold.}
\label{tab:sice}
\begin{adjustwidth}{-\extralength}{0cm}
        \begin{tabularx}{\fulllength}{cCCCCCcCc}
        \toprule
        \multirow{2}{*}{\textbf{Methods}\vspace{-4pt}} & \multicolumn{8}{c}{\textbf{SICE Dataset}} \\
        \cmidrule{2-9}
        ~& \textbf{PSNR \boldmath($\uparrow$)} & \textbf{SSIM \boldmath($\uparrow$)} & \textbf{CC \boldmath($\uparrow$)} & \textbf{AG \boldmath($\uparrow$)} & \textbf{VIF \boldmath($\uparrow$)} & \textbf{MEF-SSIM \boldmath($\uparrow$)} & \textbf{TMQI \boldmath($\uparrow$)} & \textbf{MS-SSIM \boldmath($\uparrow$)} \\
        \midrule
        Deepfuse & 17.580 & 0.883 & 0.904 & 5.775 & 1.320 & 0.843 & 0.858 & 0.886 \\
        U2Fusion & 17.670 & 0.863 & 0.921 & 4.425 & 1.037 & 0.897 & 0.837 & 0.813 \\
        FusionDN & 18.086 & 0.649 & 0.762 & 5.143 & 1.301 & 0.825 & 0.804 & 0.827 \\
        Retinex-MEF & 19.136 & 0.893 & 0.841 & 7.025 & 1.283 & 0.889 & 0.845 & 0.857 \\
        IFCNN & 19.130 & 0.894 & 0.886 & \textbf{8.459} & 1.371 & 0.896 & 0.848 & 0.862 \\
        MEFGAN & 19.710 & 0.902 & 0.927 & 6.077 & 1.201 & 0.819 & 0.865 & 0.881 \\
        MEFLUT & 21.894 & 0.807 & 0.825 & 6.894 & 1.128 & \textbf{0.985} & 0.856 & 0.875 \\
        TransMEF & 21.601 & 0.797 & 0.817 & 6.752 & 1.105 & 0.978 & 0.853 & 0.869 \\
        MEFNet & 21.583 & 0.776 & 0.710 & 7.624 & 1.142 & 0.974 & 0.849 & 0.861 \\
        DPEMEF & 19.230 & 0.904 & 0.923 & 7.168 & 1.324 & 0.916 & 0.852 & 0.905 \\
        DMEF & 19.435 & \textbf{0.999} & 0.941 & 8.216 & 1.365 & 0.924 & 0.868 & 0.912 \\
        AGAL & 22.686 & 0.952 & 0.948 & 8.392 & 1.376 & 0.981 & \textbf{0.873} & 0.918 \\
        Yin et al. & \textbf{24.412} & 0.915 & \textbf{0.951} & 8.427 & 1.086 & 0.936 & 0.870 & \textbf{0.923} \\
        HoLoCo & 21.335 & 0.939 & 0.944 & 8.330 & \textbf{1.380} & 0.921 & 0.871 & 0.901 \\
        \bottomrule
        \end{tabularx}
\end{adjustwidth}
\end{table}

Table~\ref{tab:mefb} reports results on the more comprehensive MEFB dataset. Compared with earlier models, recent approaches achieve more balanced performance across a wider range of evaluation metrics. The improvements are noticeable in perceptual quality and information preservation, reflecting more effective feature fusion and exposure modeling.

\textls[-15]{To evaluate existing deghosting algorithms, quantitative comparisons are conducted on the Kalantari, Hu, and Tel datasets using PSNR-$\mu$, SSIM-$\mu$, PSNR-L, SSIM-L, and HDR-VDP-2.}

Table~\ref{tab:kalantari_hu} shows that learning-based methods outperform traditional approaches on both Kalantari and Hu datasets. Early methods such as Hu and Sen achieve limited results, while CNN-based models provide clear improvements. Recent approaches incorporating attention mechanisms, GANs, or Transformers further enhance reconstruction fidelity and perceptual quality. Table~\ref{tab:tel} reports results on the more challenging Tel dataset, with the best values in bold. Learning-based methods show stronger robustness in dynamic scenes and consistently perform better across most metrics, reflecting improved handling of motion and exposure variations.

\startlandscape

\begin{table}[H]
\caption{Quantitative comparison of MEF algorithms on the MEFB dataset. $\uparrow$ means higher is better, while $\downarrow$ means lower is better. The best results are highlighted in bold.}
\label{tab:mefb}
        \begin{tabularx}{\textwidth}{cCCCccCcCCCcCcC}
        \toprule
        \multirow{2}{*}{\textbf{Methods\vspace{-4pt}}} & \multicolumn{14}{c}{\textbf{MEFB Dataset}} \\
        \cmidrule{2-15}
        ~& \textbf{EN \boldmath($\uparrow$)} & \textbf{FMI \boldmath($\uparrow$)} & \textbf{NMI \boldmath($\uparrow$)} & \textbf{PSNR \boldmath($\uparrow$)} & \textbf{QNCIE \boldmath($\uparrow$)} & \textbf{AG \boldmath($\uparrow$)} & \boldmath$Q^{AB/F}$ ($\uparrow$) & \textbf{CE \boldmath($\downarrow$)} & \boldmath$Q_P$ ($\uparrow$) & \boldmath$Q_W$ ($\uparrow$) & \textbf{MEF-SSIM \boldmath($\uparrow$)} & \boldmath$Q_{CB}$ ($\uparrow$) & \boldmath$Q_{CV}$ ($\downarrow$) & \textbf{VIF \boldmath($\uparrow$)} \\
        \midrule
        DeepFuse    & 6.8504 & 0.8727 & 0.7408 & 57.1035 & 0.8177 & 3.4920 & 0.3884 & 3.0852 & 0.3517 & 0.5478 & 0.8968 & 0.3892 & 362.9800 & 0.5114 \\
        MEF-GAN     & 6.9547 & 0.8456 & 0.5727 & 56.9474 & 0.8132 & 4.6702 & 0.2836 & 2.8222 & 0.1239 & 0.3002 & 0.7722 & 0.3844 & 618.6932 & 0.5810 \\
        EMEF        & 7.2195 & 0.8545 & 0.6114 & 53.6245 & 0.8141 & \textbf{6.9694} & \textbf{0.6933} & \textbf{1.7607} & \textbf{0.7254} & \textbf{0.8853} & 0.8751 & 0.3969 & 312.8924 & 0.7842 \\
        FusionDN    & 7.3293 & 0.8770 & 0.7251 & 56.9770 & 0.8178 & 6.7934 & 0.5363 & 2.9357 & 0.5044 & 0.7761 & 0.9240 & 0.4386 & 325.1348 & \textbf{0.9363} \\
        MSCA-MEF    & 6.8604 & 0.8856 & 0.8236 & 57.1259 & \textbf{0.8238} & 4.5950 & 0.6031 & 2.8039 & 0.5601 & 0.8062 & 0.9509 & 0.4004 & 251.2233 & 0.7651 \\
        U2Fusion    & 6.7392 & 0.8821 & 0.7675 & 57.0550 & 0.8179 & 5.5829 & 0.5356 & 2.9761 & 0.5046 & 0.7874 & 0.9304 & 0.4174 & 253.7540 & 0.8358 \\
        DPE-MEF     & 7.2383 & 0.8788 & 0.6120 & 57.1051 & 0.8141 & 6.6607 & 0.5995 & 4.1311 & 0.5612 & 0.8304 & 0.9452 & 0.3942 & 257.3125 & 0.7885 \\
        MEFNet      & \textbf{7.3899} & 0.8896 & 0.5967 & 56.5941 & 0.8166 & 6.0104 & 0.6746 & 3.0300 & 0.5954 & 0.8655 & 0.9139 & \textbf{0.4816} & 593.4327 & 0.8470 \\
        IFCNN       & 7.0347 & 0.8824 & 0.7708 & \textbf{57.1951} & 0.8186 & 6.0123 & 0.5960 & 3.4098 & 0.5616 & 0.8336 & 0.9432 & 0.4112 & \textbf{247.7693} & 0.7016 \\
        TransMEF    & 6.8603 & \textbf{0.8910} & \textbf{0.9229} & 57.1319 & 0.8237 & 4.5949 & 0.6035 & 2.8038 & 0.5649 & 0.8059 & 0.9499 & 0.4001 & 253.3766 & 0.7658 \\
        FFMEF       & 6.9942 & 0.8880 & 0.8311 & 57.1918 & 0.8206 & 5.0976 & 0.6584 & 2.7933 & 0.6073 & 0.8357 & \textbf{0.9621} & 0.4102 & 248.0949 & 0.7119 \\
        \bottomrule
        \end{tabularx}
\end{table}
\vspace{-12pt}

\begin{table}[H]
\caption{Quantitative comparison of deghosting algorithms on the Kalantari and Hu datasets. $\uparrow$ means higher is better, while $\downarrow$ means lower is better. The best results are highlighted in bold.}
\label{tab:kalantari_hu}
        \begin{tabularx}{\textwidth}{cCCCCcCCCCc}
        \toprule
        \multirow{2}{*}{\textbf{Methods\vspace{-4pt}}} & \multicolumn{5}{c}{\textbf{Kalantari Dataset}} & \multicolumn{5}{c}{\textbf{Hu Dataset}} \\
        \cmidrule{2-11}
        ~& \textbf{PSNR-\boldmath$\mu$ ($\uparrow$)} & \textbf{SSIM-\boldmath$\mu$ ($\uparrow$)} & \textbf{PSNR-L \boldmath($\uparrow$)} & \textbf{SSIM-L \boldmath($\uparrow$)} & \textbf{HDR-VDP-2 \boldmath($\uparrow$)} 
        & \textbf{PSNR-\boldmath$\mu$ ($\uparrow$)} & \textbf{SSIM-\boldmath$\mu$ ($\uparrow$)} & \textbf{PSNR-L \boldmath($\uparrow$)} & \textbf{SSIM-L \boldmath($\uparrow$)} & \textbf{HDR-VDP-2 \boldmath($\uparrow$)} \\
        \midrule
        Hu \cite{hu2013hdr}  & 32.19 & 0.9716 & 30.84 & 0.9506 & 55.25 & 36.56 & 0.9824 & 36.94 & 0.9877 & 67.58 \\
        Sen \cite{sen2012robust}  & 40.95 & 0.9832 & 38.31 & 0.9753 & 60.33 & 31.48 & 0.9531 & 33.58 & 0.9634 & 66.39 \\
        DeepHDR \cite{wu2018deep} & 41.62 & 0.9865 & 40.88 & 0.9858 & 57.37 & 44.70 & 0.9945 & 44.27 & 0.9960 & 68.90 \\
        Kalantari \cite{kalantari2017deep}  & 42.74 & 0.9877 & 40.72 & 0.9824 & 62.87 & 41.60 & 0.9914 & 43.76 & 0.9938 & 64.70 \\
        NHDRRNet \cite{yan2020deep} & 42.41 & 0.9887 & 41.08 & 0.9861 & 61.21 & 45.15 & 0.9956 & 48.75 & 0.9981 & 74.86 \\
        Chung et al. \cite{chung2022high}  & 43.65 & 0.9894 & 41.67 & 0.9867 & 64.46 & 43.77 & 0.9930 & 46.31 & 0.9975 & 67.82 \\
        AHDRNet \cite{yan2019attention}  & 43.62 & 0.9900 & 41.03 & 0.9862 & 62.30 & 45.76 & 0.9956 & 49.22 & 0.9980 & 75.04 \\
        HDR-GAN \cite{niu2021hdr} & 43.92 & 0.9905 & 41.57 & 0.9865 & 65.45 & 45.86 & 0.9945 & 49.14 & 0.9989 & 75.19 \\
        DiffHDR \cite{yan2023toward} & 44.11 & 0.9911 & 41.73 & 0.9885 & 65.52 & 48.03 & 0.9954 & 50.23 & 0.9989 & 76.22 \\
        HDR-Transformer \cite{liu2022ghost} & 44.32 & 0.9916 & 42.18 & 0.9884 & 64.63 & 46.14 & 0.9961 & 50.04 & 0.9988 & 68.92 \\
        SCTNet \cite{tel2023alignment} & 44.43 & 0.9918 & 42.21 & 0.9891 & 66.64 & 48.10 & 0.9963 & 51.14 & 0.9991 & 77.14 \\
        RFG-HDR \cite{lee2024rfg} & 44.21 & 0.9915 & 42.16 & 0.9893 & 66.47 & - & - & - & - & - \\
        HyHDRNet \cite{yan2023unified} & 44.64 & 0.9915 & 42.47 & 0.9894 & 66.05 & 48.46 & 0.9959 & 51.91 & 0.9991 & 77.24 \\
        SAFNet \cite{kong2024safnet} & 44.66 & 0.9919 & \textbf{43.18} & 0.9901 & 66.69 & - & - & - & - & - \\
        LFDiff \cite{hu2024generating} & 44.76 & 0.9919 & 42.59 & \textbf{0.9906} & 66.54 & 48.74 & \textbf{0.9968} & 52.10 & \textbf{0.9993} & 77.35 \\
        AFUNet \cite{li2025afunet} & \textbf{44.91} & \textbf{0.9923} & 42.59 & \textbf{0.9906} & \textbf{66.75} & \textbf{48.83} & \textbf{0.9968} & \textbf{52.13} & 0.9991 & \textbf{77.44} \\
        \bottomrule
        \end{tabularx}
\end{table}

\finishlandscape

\begin{table}[H]
\caption{Quantitative comparison of deghosting algorithms on the Tel dataset. $\uparrow$ means higher is better, while $\downarrow$ means lower is better. The best results are highlighted in bold.}
\label{tab:tel}

\begin{adjustwidth}{-\extralength}{0cm}
\begin{tabularx}{\fulllength}{cCCCCC}
\toprule
\multirow{2}{*}{\textbf{Methods\vspace{-4pt}}} & \multicolumn{5}{c}{\textbf{Tel Dataset}} \\
\cmidrule{2-6}
& \textbf{PSNR-\boldmath$\mu$ ($\uparrow$)} & \textbf{SSIM-\boldmath$\mu$ ($\uparrow$)} & \textbf{PSNR-L \boldmath($\uparrow$)} & \textbf{SSIM-L \boldmath($\uparrow$)} & \textbf{HDR-VDP-2 \boldmath($\uparrow$)} \\
\midrule
NHDRRNet \cite{yan2020deep} & 36.68 & 0.9590 & 39.61 & 0.9853 & 65.41 \\
DeepHDR \cite{wu2018deep} & 40.05 & 0.9794 & 43.37 & 0.9924 & 67.09 \\
AHDRNet \cite{yan2019attention} & 42.08 & 0.9837 & 45.30 & 0.9943 & 68.80 \\
HDR-GAN \cite{niu2021hdr} & 41.71 & 0.9832 & 44.87 & 0.9949 & 69.57 \\
HDR-Transformer \cite{liu2022ghost} & 42.39 & 0.9844 & 46.35 & 0.9948 & 69.23 \\
DiffHDR \cite{yan2023toward} & 42.18 & 0.9841 & 45.63 & 0.9946 & 69.88 \\
SAFNet \cite{kong2024safnet} & 42.21 & 0.9852 & 47.73 & 0.9953 & 68.99 \\
SCTNet \cite{tel2023alignment} & 42.55 & 0.9850 & 47.51 & 0.9952 & 70.66 \\
AFUNet \cite{li2025afunet} & \textbf{43.31} & \textbf{0.9876} & \textbf{47.83} & \textbf{0.9959} & \textbf{71.08} \\
\bottomrule
\end{tabularx}
\end{adjustwidth}
\end{table}

\subsection{Subjective Comparison}
This subsection conducts a systematic subjective evaluation on existing MEF and deghosting algorithms. All compared methods are selected from representative works with publicly available code and pretrained models, and are evaluated under the same input conditions to ensure fairness and reproducibility. In addition, all results are processed using the tonemap \cite{zhang2023self} function in MATLAB to maintain a consistent visual presentation.

\begin{figure}[H]

\begin{adjustwidth}{-\extralength}{0cm}
\centering 
\includegraphics[width=.89\linewidth]{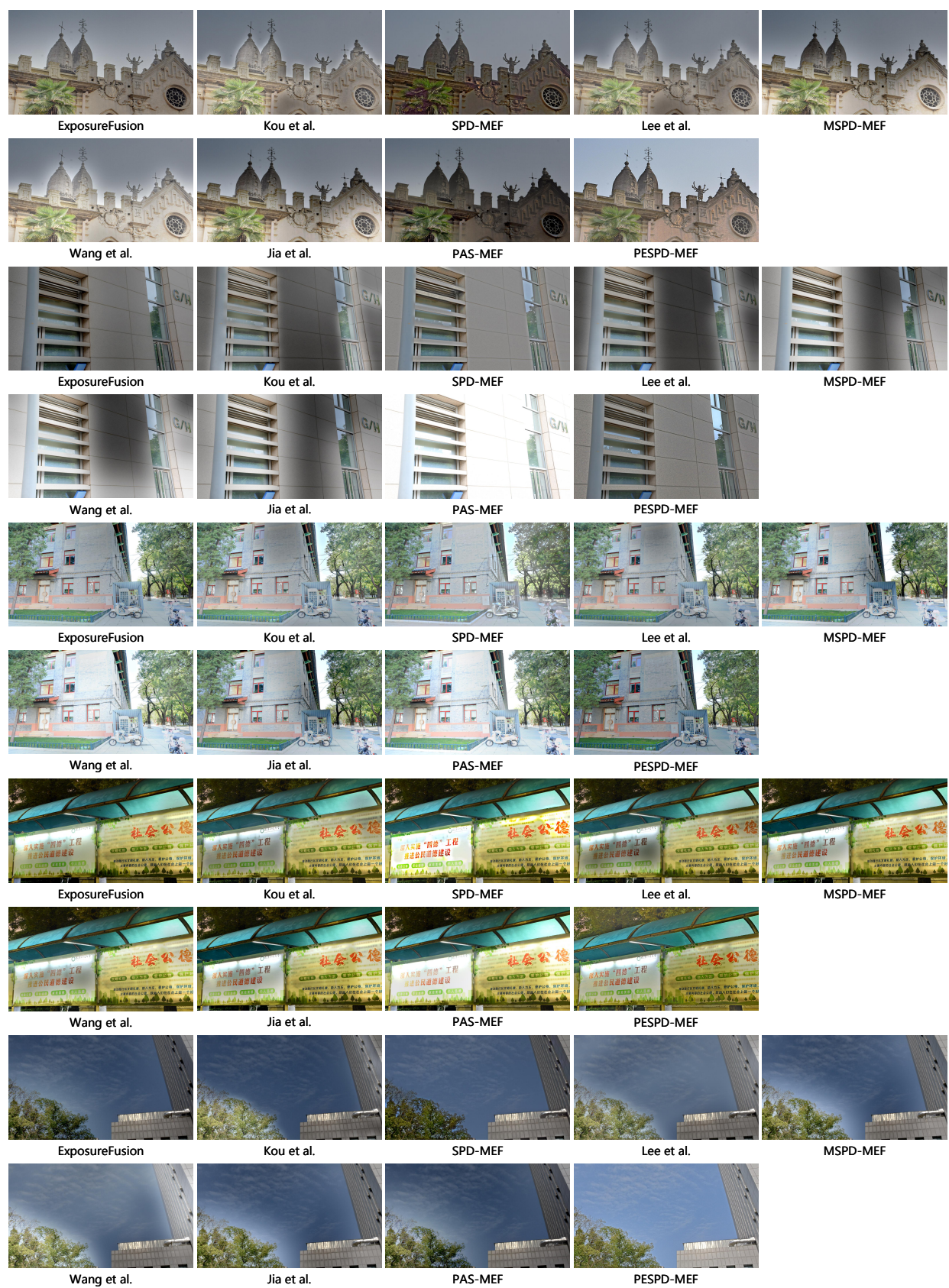}
\end{adjustwidth}
\caption{Results of traditional SOTA MEF algorithms. Representative methods include ExposureFusion~\cite{mertens2007exposure}, SPD-MEF~\cite{ma2017robust}, MSPD-MEF~\cite{li2020fast}, Wang et al.~\cite{wang2019detail}, PAS-MEF~\cite{karakaya2022pas}, Jia et al.~\cite{jia2022multi}, Kou et al.~\cite{kou2015gradient}, Lee et al.~\cite{lee2018multi}, and PESPD-MEF~\cite{zhang2023multi}.}
\label{fig:tradition2}
\end{figure}

\begin{figure}[H]

\begin{adjustwidth}{-\extralength}{0cm}
\centering 
\includegraphics[width=.89\linewidth]{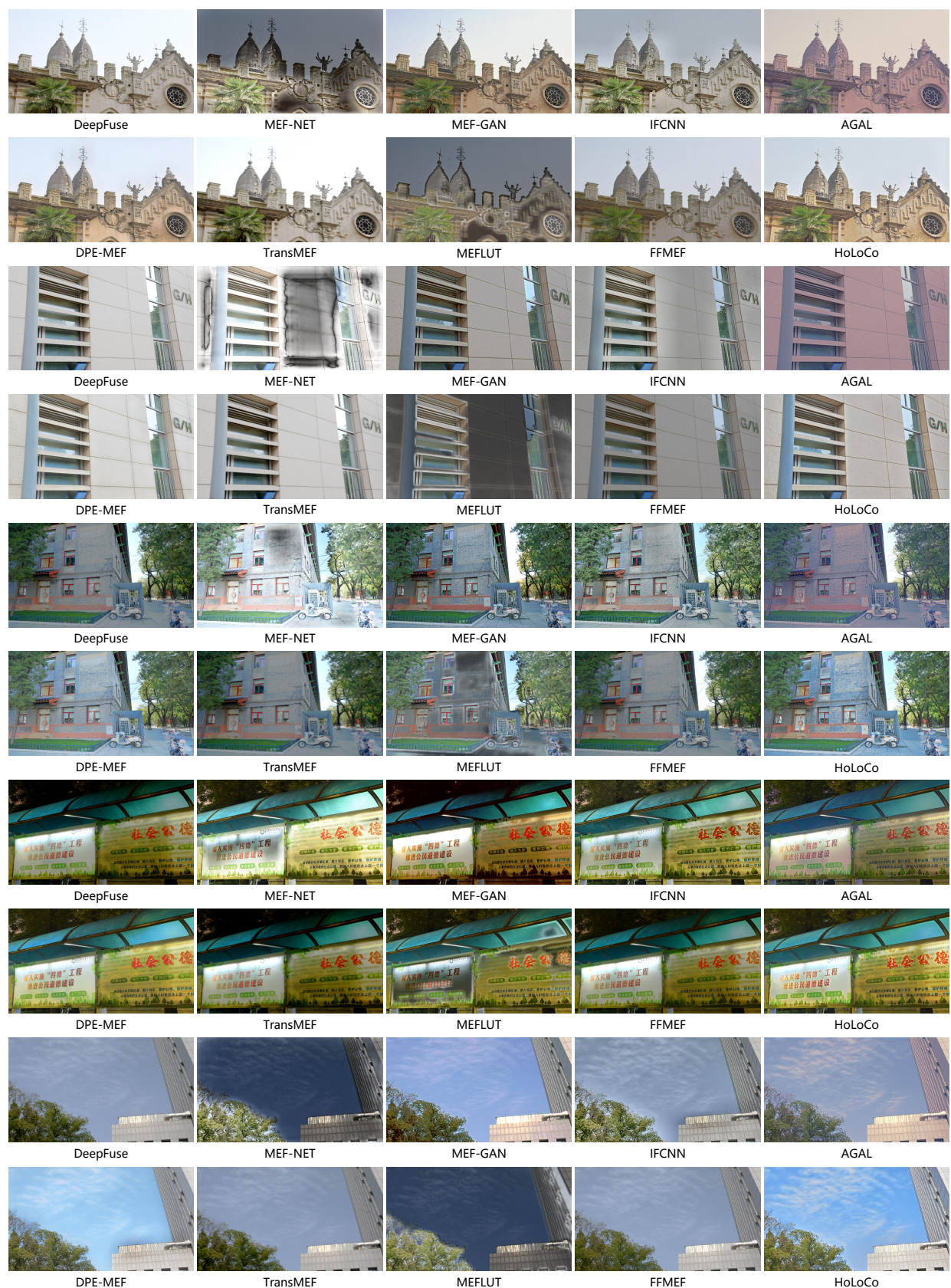}
\end{adjustwidth}
\caption{Results of representative deep learning-based MEF algorithms. Representative methods include DeepFuse~\cite{ram2017deepfuse}, MEF-NET~\cite{1ma2019}, MEF-GAN~\cite{xu2020mef}, IFCNN~\cite{zhang2020ifcnn}, AGAL~\cite{liu2022attention}, DPE-MEF~\cite{han2022multi}, TransMEF~\cite{qu2022transmef}, MEFLUT~\cite{jiang2023meflut}, FFMEF~\cite{zheng2023efficient}, and HoLoCo~\cite{liu2023holoco}.}
\label{fig:Brightness1}
\end{figure}

\begin{figure}[H]

\begin{adjustwidth}{-\extralength}{0cm}
\centering 
\includegraphics[width=.89\linewidth]{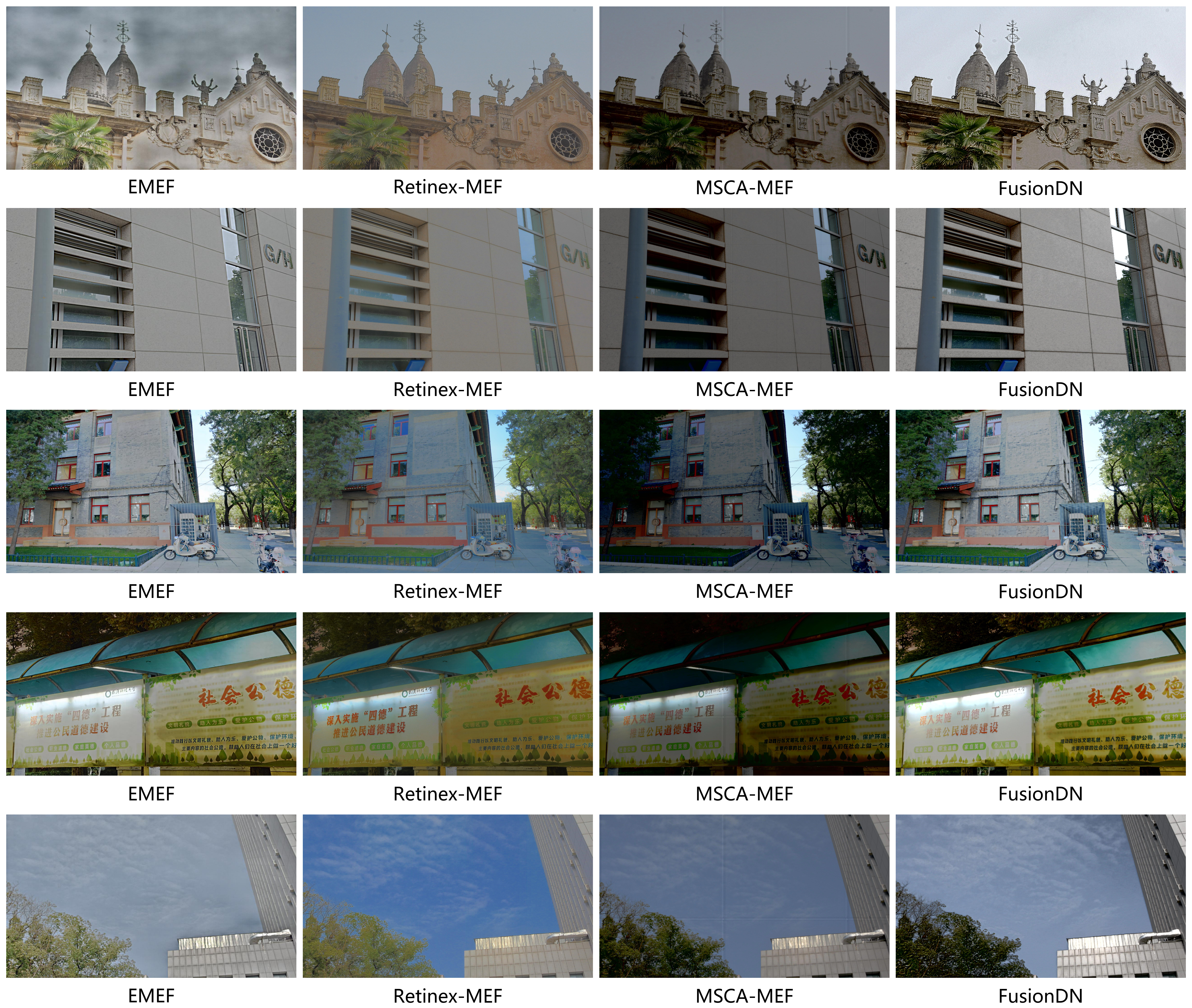}
\end{adjustwidth}
\caption{Results of representative deep learning-based MEF algorithms. Representative methods include FusionDN~\cite{xu2020fusiondn}, EMEF~\cite{liu2023emef}, MSCA-MEF~\cite{liu2023multi}, and Retinex-MEF~\cite{bai2025retinex}.}
\label{fig:Brightness2}
\end{figure}

Five representative scenes are selected for comparison of MEF algorithms, as illustrated in Figures \ref{fig:tradition2}--\ref{fig:Brightness2}. The first three scenes are from the SICE dataset \cite{cai2018learning}, which covers diverse exposure conditions and complex illumination variations. The fourth scene is from the Canon5D4 dataset \cite{xu2021deep}, designed to evaluate performance under real camera acquisition settings. The fifth scene corresponds to a real-world captured example, providing additional validation in practical scenarios.

These qualitative results in the figures  
provide a comprehensive comparison between traditional SOTA methods and deep learning-based approaches regarding brightness consistency. While the traditional EPS-based method in \cite{1kou2017} indeed preserves fine details of highlight and shadow regions in the fused image better than the MEF algorithm in~\cite{mertens2007exposure}, the EPS-based MEF algorithms suffer from halos, for example, on the edges of the building in the first example. The BOR artifacts are an issue for all exposure fusion algorithms in pixel space if the set $\Omega_f$ only includes two LER images. In contrast, in \mbox{Figures \ref{fig:Brightness1} and \ref{fig:Brightness2}}, feature-space deep learning methods (AGAL \cite{liu2022attention}, HoLoCo \cite{liu2023holoco}, and Retinex-MEF \cite{bai2025retinex}) can address both BOR artifacts and halos. However, the global contrast or depth of the HDR scene might not be well preserved, and the fused image could look like a flat ``cartoony'' rendition. In addition, these models still exhibit slight over-enhancement or color deviation under extreme exposure differences, as shown in the third and fourth cases of  Figure \ref{fig:Brightness1}, where AGAL \cite{liu2022attention} and HoLoCo \cite{liu2023holoco} had evident color shift and limited color depth. One more issue for almost all existing MEF algorithms is that the fused image approaches the image in the set $\Omega_f$. It is impossible to preserve fine details of highlight and shadow regions in the fused image if these details are not captured by the set $\Omega_f$. These observations highlight the ongoing challenges in achieving high luminance consistency.

Therefore, it is still desirable to study more real-world adaptive MEF even though there are many SOTA MEF algorithms. We argue that the fused image should approach the real-world HDR scene rather than the set $\Omega_f$. Despite achieving higher visual quality, these approaches above primarily approximate the LDR distribution of their training datasets rather than the true physical dynamic range of complex real-world scenes. Their performance is heavily biased toward the aesthetic priors of curated ground truths, often failing to recover reliable radiance under extreme illumination. This suggests that existing methods are limited by data-driven priors, falling short of a true physical restoration of a full scene’s high dynamic range. This is still an undiscovered problem for MEF algorithms.

\begin{figure}[H]

\begin{adjustwidth}{-\extralength}{0cm}
\centering 
\includegraphics[width=.89\linewidth]{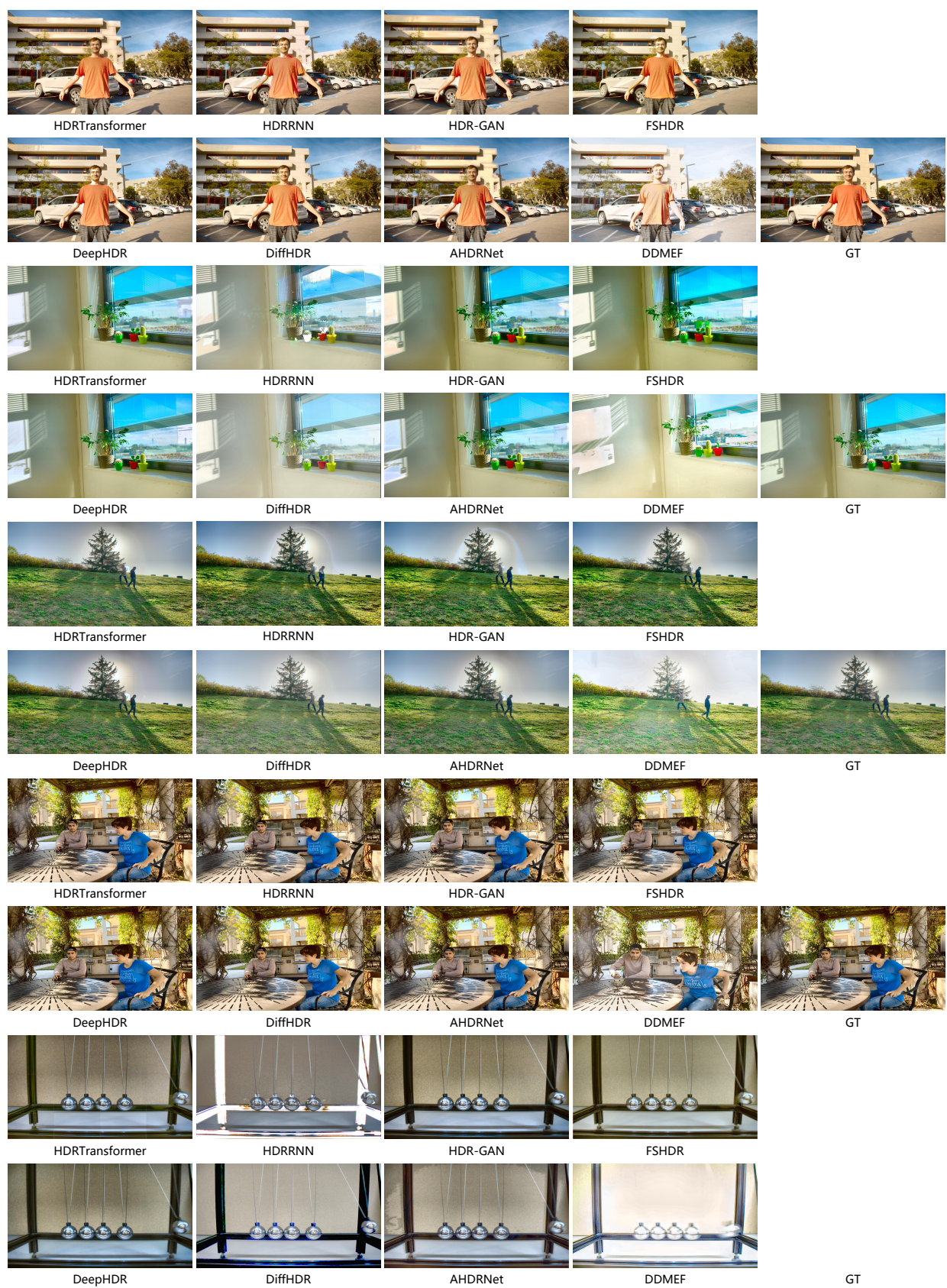}
\end{adjustwidth}
\caption{Results of representative deep learning-based deghosting algorithms. Representative methods include DeepHDR~\cite{wu2018deep}, AHDRNet~\cite{yan2019attention}, FSHDR~\cite{prabhakar2021labeled}, HDRRNN~\cite{prabhakar2021self}, HDR-GAN~\cite{niu2021hdr}, HDRTransformer~\cite{liu2022ghost}, DDMEF~\cite{tan2023deep}, and DiffHDR~\cite{yan2023toward}.}
\label{fig:crop_Deghost2}
\end{figure}

\begin{figure}[H]

\begin{adjustwidth}{-\extralength}{0cm}
\centering 
\includegraphics[width=0.89\linewidth]{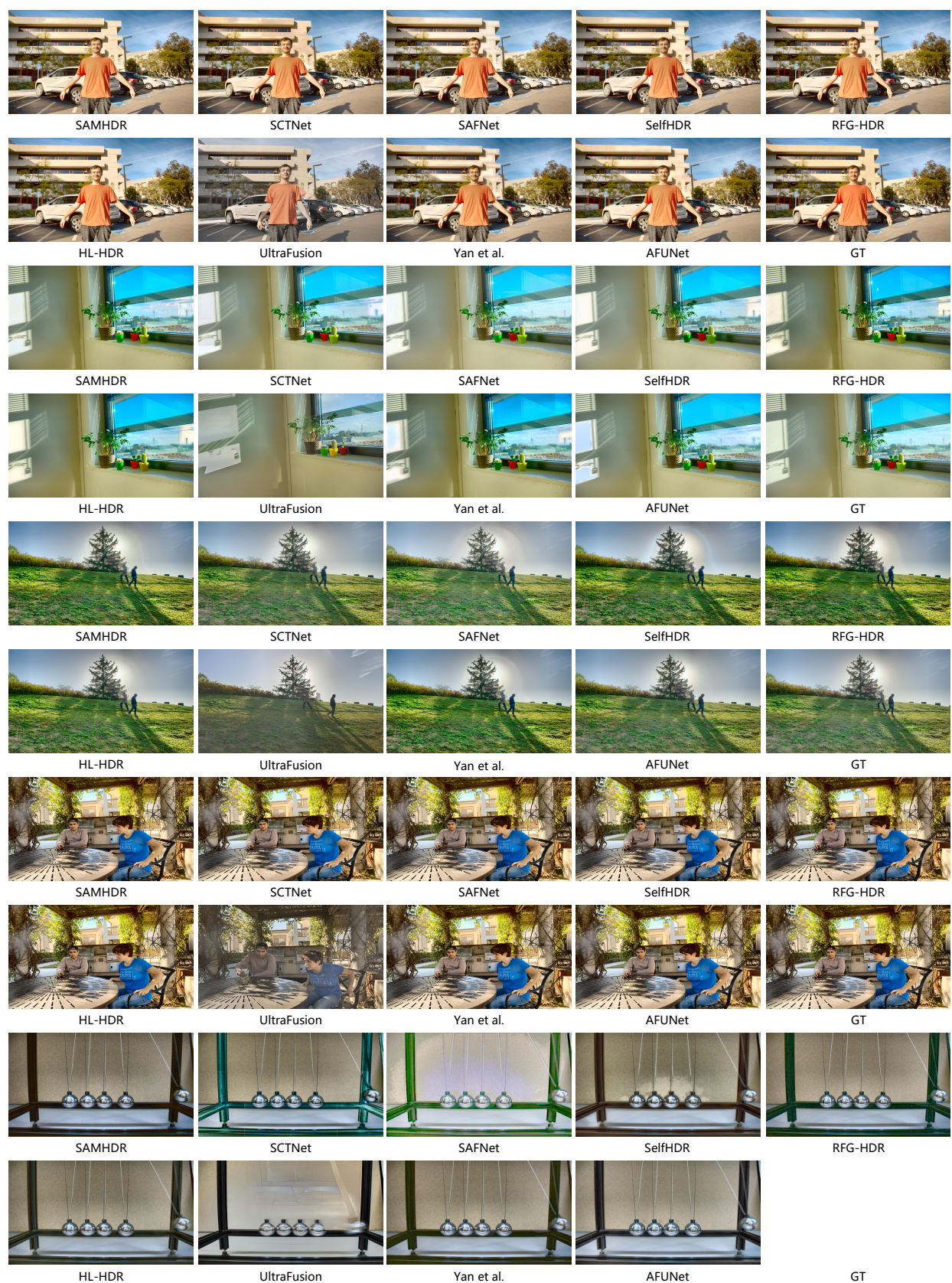}
\end{adjustwidth}
\caption{Results of representative deep learning-based deghosting algorithms. Representative methods include SAMHDR~\cite{li2024single}, SCTNet~\cite{tel2023alignment}, SAFNet~\cite{kong2024safnet}, RFG-HDR~\cite{lee2024rfg}, HL-HDR~\cite{zhang2024hl}, UltraFusion~\cite{chen2025ultrafusion}, Yan et al.~\cite{yan2024dynamic}, SelfHDR~\cite{zhang2023self}, and AFUNet~\cite{li2025afunet}.}
\label{fig:crop_Deghost1}
\end{figure}

For the deghosting task in dynamic scenes, three representative data sources are considered, as shown in Figures \ref{fig:crop_Deghost2} and \ref{fig:crop_Deghost1}. The first and fourth groups are taken from the Kalantari dataset \cite{kalantari2017deep}, which provides ground-truth references and is widely used for evaluating dynamic HDR reconstruction. The second and third groups come from the TEl dataset \cite{tel2023alignment}, which contains scenes with different levels of motion. The fifth group is from the Tursun \cite{tursun2016objective} dataset, which only provides test samples and is used to examine the generalization ability of different methods.

 These visual comparisons demonstrate that feature-space approaches, particularly those utilizing attention mechanisms or Transformer architectures like  AHDRNet \cite{yan2019attention}, HDR-Transformer \cite{liu2022ghost} and SCTNet \cite{tel2023alignment}, exhibit significantly higher robustness in handling dynamic regions compared to early CNN-based models. By suppressing motion-induced artifacts, these methods preserve structural integrity and produce visually consistent results even under complex non-rigid motion, as shown by the hands in the first case. Although most models maintain reasonable performance and sharp textures in unseen data distributions, such as the Tursun \cite{tursun2016objective} dataset, slight blurring or local inconsistencies can still be observed in extremely challenging regions where matching becomes difficult, as shown by the curved strings in the last case, which should be straight in the real world.

\begin{figure}[H]
   {\captionsetup{position=bottom,justification=centering} \subfloat[Input 1]{\includegraphics[width=0.25\textwidth]{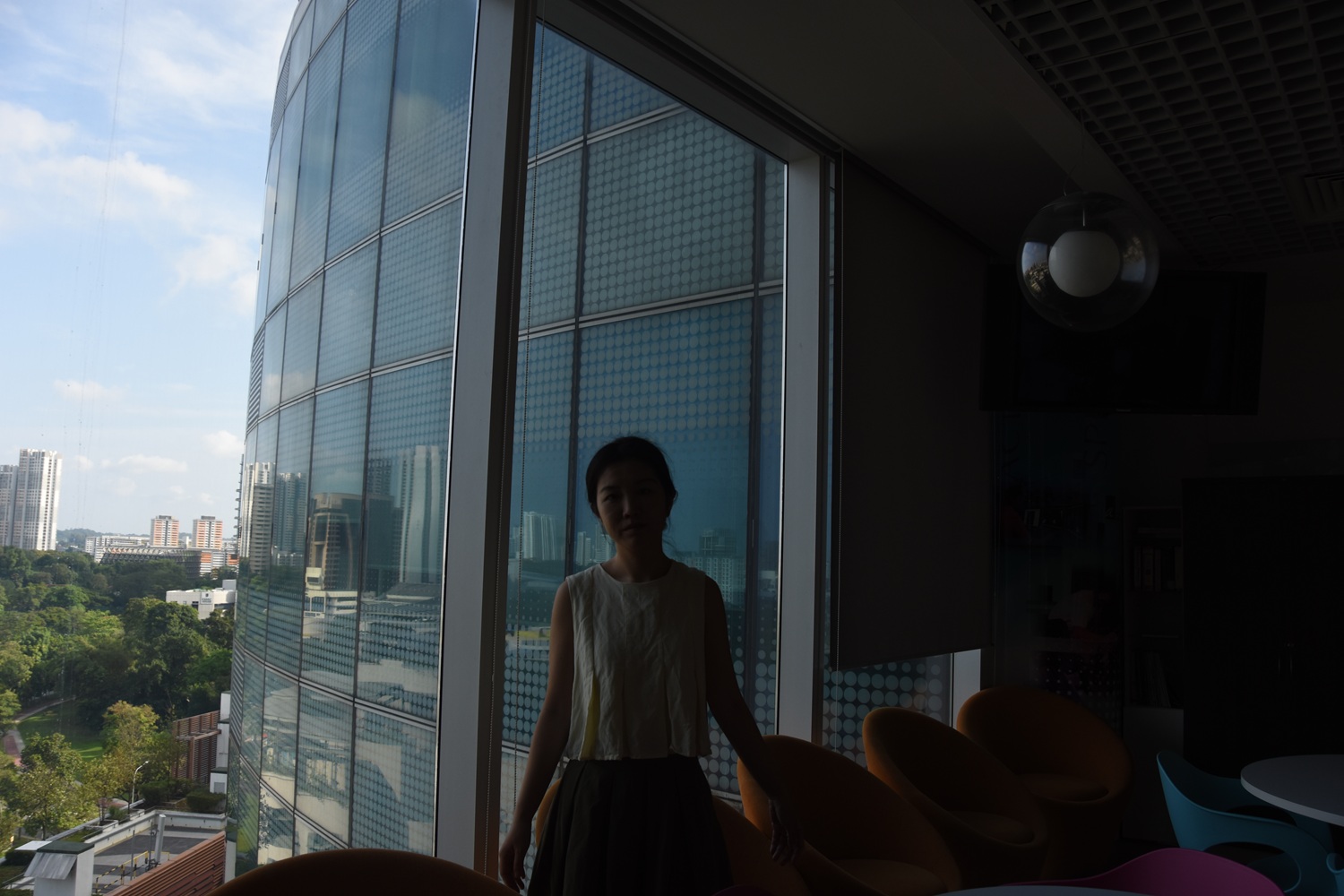}}
    \hfill
    \subfloat[Input 2]{\includegraphics[width=0.25\textwidth]{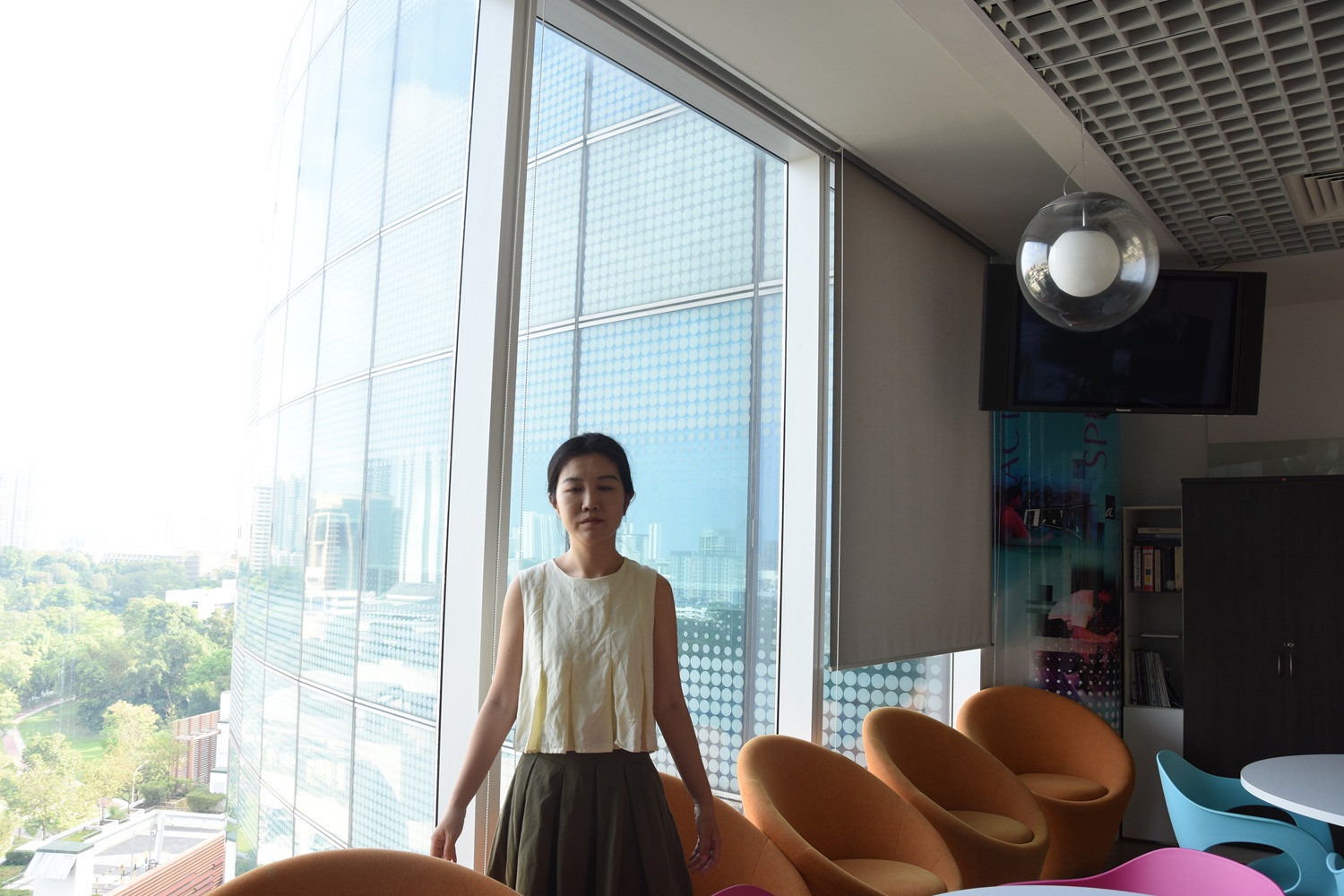}}
    \hfill
    \subfloat[Input 3]{\includegraphics[width=0.25\textwidth]{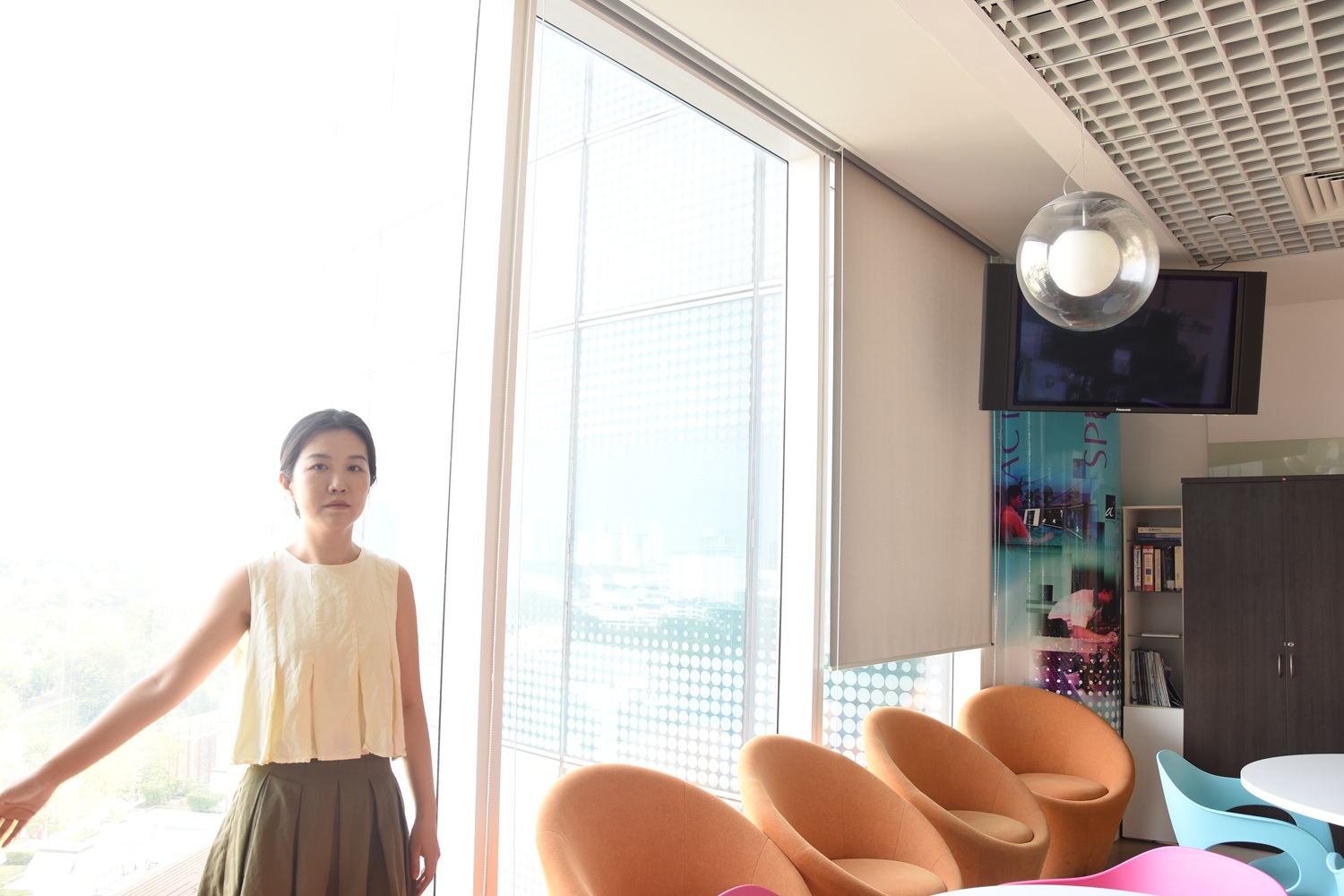}}
     \hfill
    \subfloat[UltraFusion Result]{\includegraphics[width=0.25\textwidth]{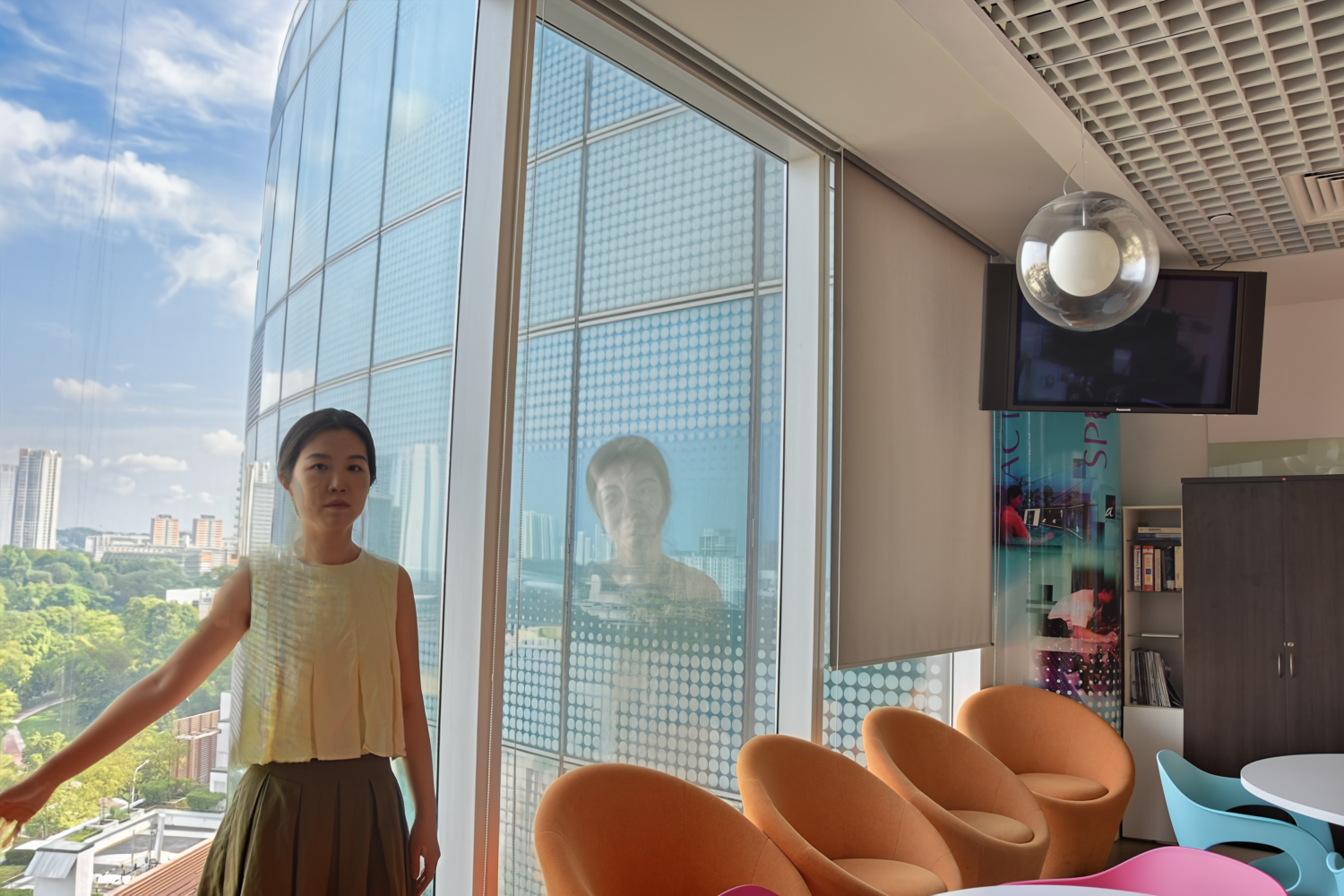}}
      \hfill
    \subfloat[DDMEF Result]{\includegraphics[width=0.25\textwidth]{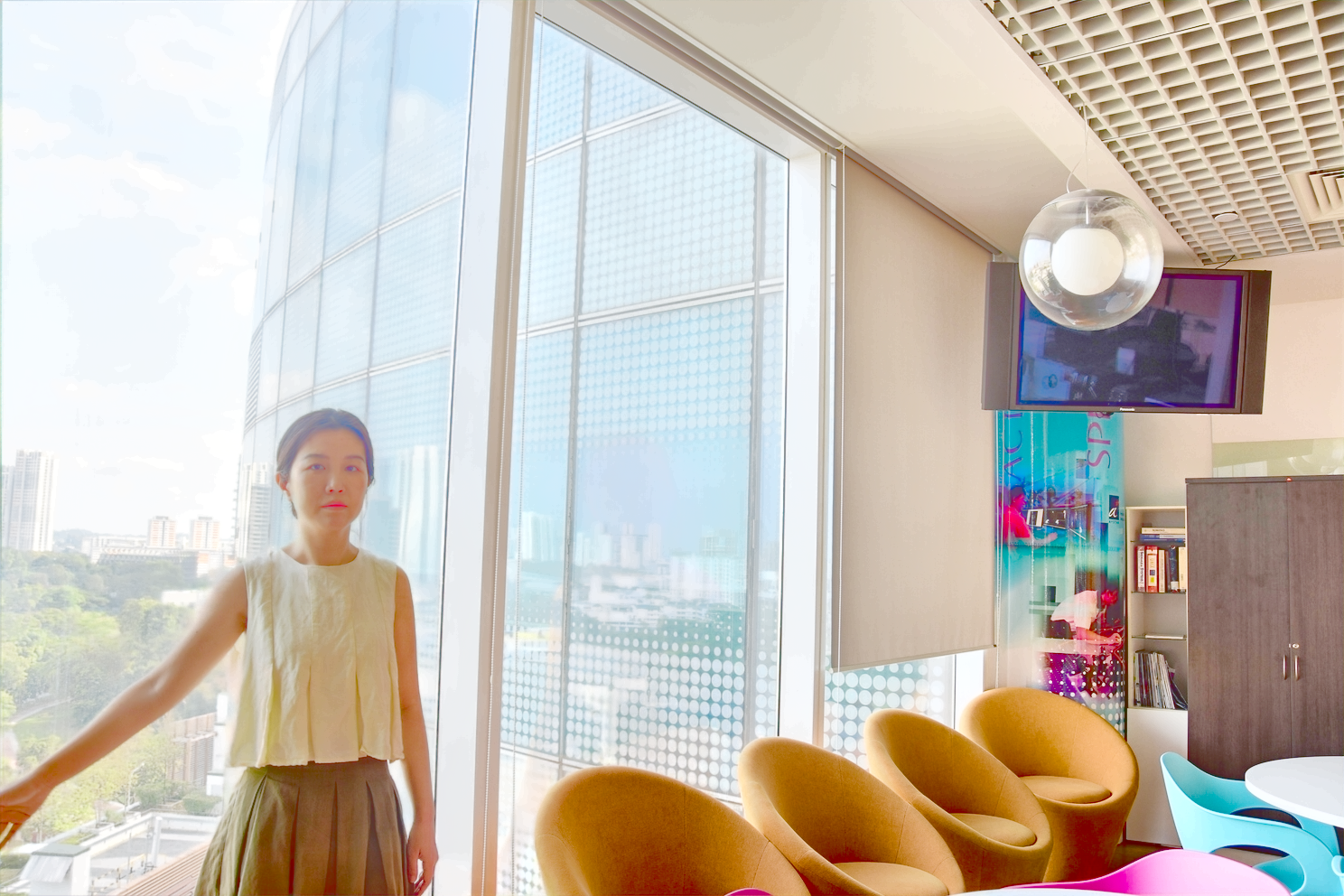}}
     \hfill
    \subfloat[AFUNet Result]{\includegraphics[width=0.25\textwidth]{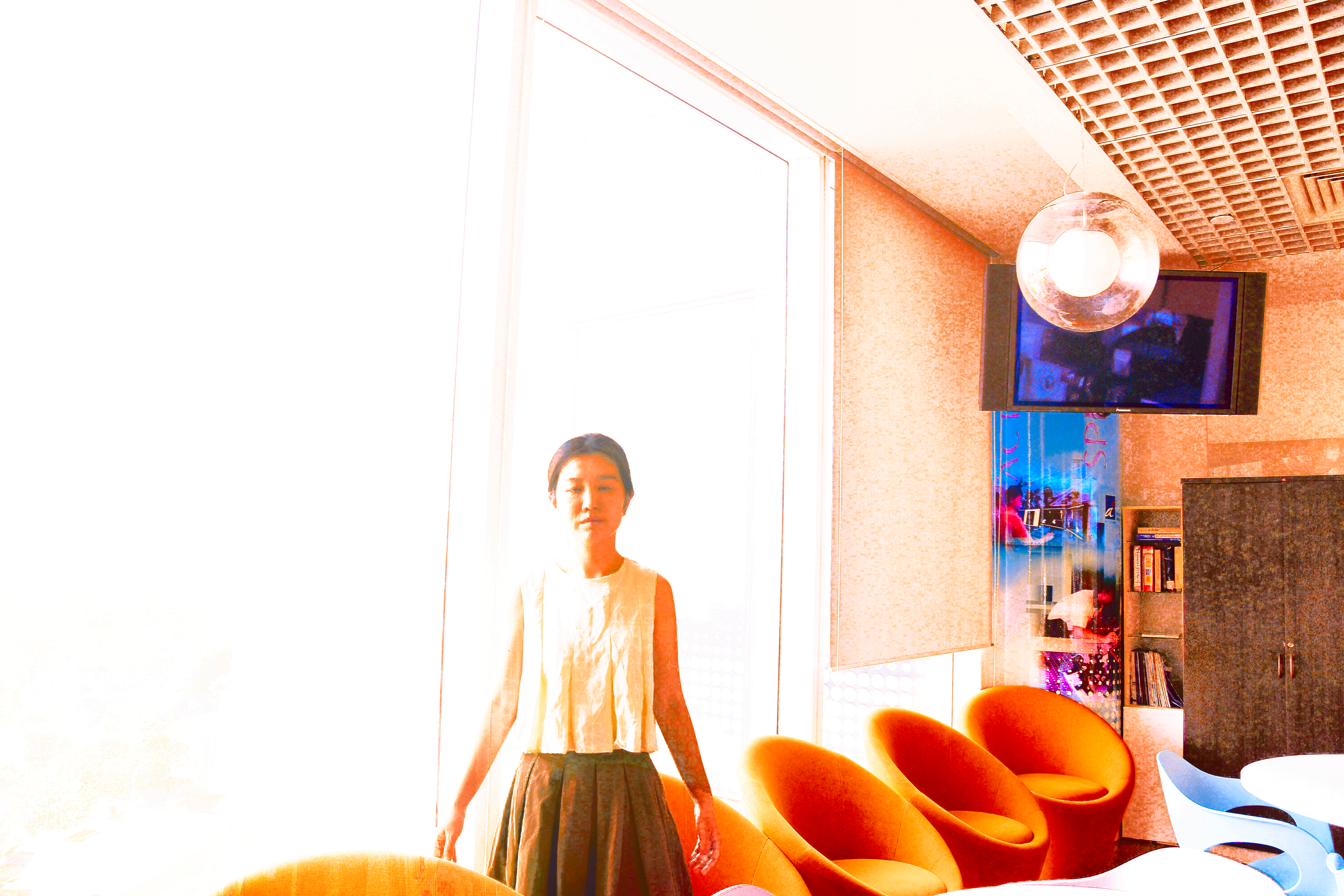}}
     \hfill
    \subfloat[SCTNet Result]{\includegraphics[width=0.25\textwidth]{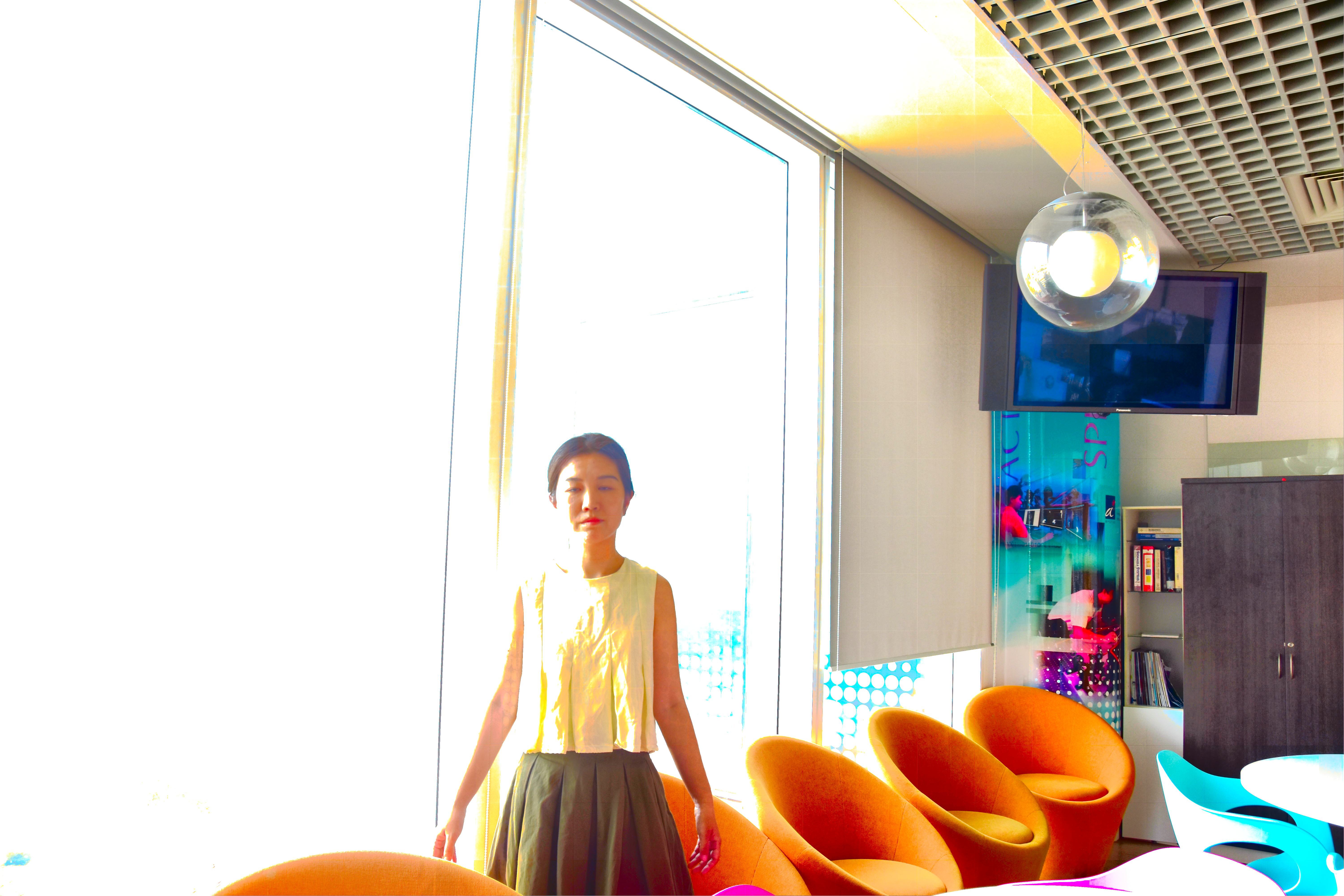}}
     \hfill
    \subfloat[FSHDR Result]{\includegraphics[width=0.25\textwidth]{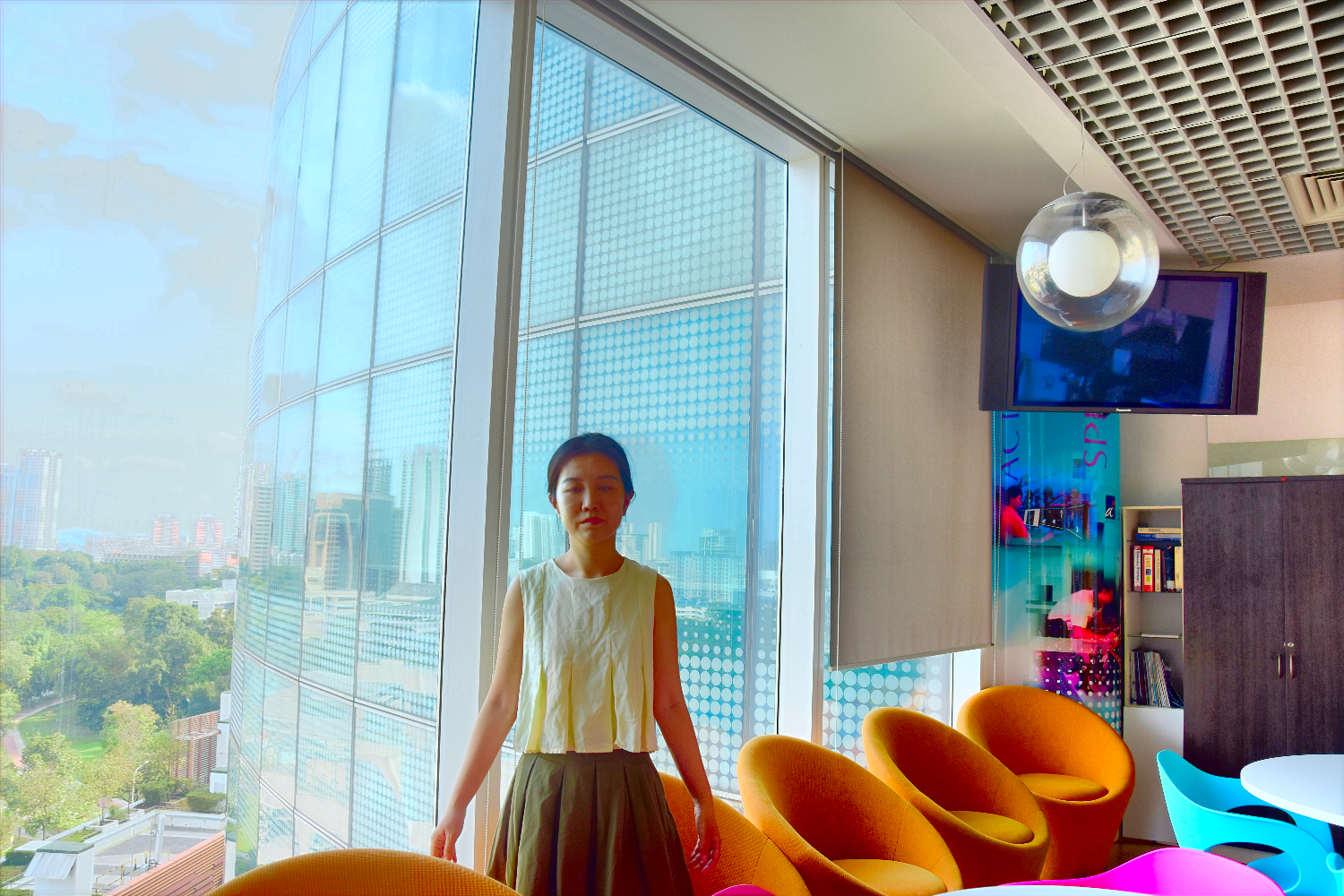}}}
    \caption{One challenging scenario for SOTA deghosting removal algorithms, including AFUNet~\cite{li2025afunet}, DDMEF \cite{tan2023deep}, SCTNet \cite{tel2023alignment}, FSHDR \cite{prabhakar2021labeled} and UltraFusion \cite{chen2025ultrafusion}. }
    \label{failcase}
\end{figure}

Despite the progress of deghosting shown in these cases, a significant gap remains between laboratory benchmarks and real-world application, primarily due to the limitations of existing datasets. Existing multi-exposure  deghosting datasets focus largely on moderate conditions and lack a sufficient representation of sequences featuring both large exposure ratios and complex object-movement-caused occlusions as shown in Figure \ref{failcase}. In such challenging scenarios, large overexposed 
or underexposed regions in the reference frame are often simultaneously occluded by moving objects in other exposures, making reliable feature matching or evidence selection extremely difficult. This combination of extreme radiance variations and non-rigid motion frequently causes existing methods to fail, leading to structural distortions or artifacts.


\vspace{-6pt}

\section{New Perspectives from Multi-Shot to Single-Shot}
Existing ghost removal algorithms fail in the following two challenging scenarios: (1) the background of the reference image includes large and heterogeneous over/underexposed regions while the corresponding regions in the other image are occluded by moving objects \cite{1zhengj2013}; and (2) the reference image includes large and heterogeneous over/underexposed regions with non-rigid movements.  One example is illustrated in Figure \ref{failcase}.
Thus, ghost removal is still an open problem for multi-shot HDR imaging. In this survey, single-shot HDR imaging is not marked as a complete replacement for multi-shot HDR imaging, nor as a parallel branch within the above multi-shot taxonomy. Instead, it is discussed as a complementary and promising future direction motivated by the unresolved multi-shot deghosting.
Single-shot HDR imaging is effective for avoiding ghosting artifacts. 
The main advantage is that HDR information is captured within one exposure event or one sensor readout process, so temporal misalignment, cross-frame occlusion, and motion-induced ghosting can be avoided at the imaging stage. However, this advantage introduces new technical challenges, including spatially varying exposure design, sensor sampling, raw-domain reconstruction, demosaicing, exposure normalization, and so on. Row-wise~\cite{1guj2010} and block-wise \cite{1hirata2021} exposures are two types of single-shot HDR imaging. The row-wise auto-exposure (AE) algorithm in \cite{1guj2010} indeed captures more  details in highlight and shadow regions of HDR scenes than conventional AE algorithms \cite{1onzon2021,1tedla2023,1lee2024,1xu2025}. However, the temporary image  is also an sRGB image. The sRGB  image has limited measurement capability of scene radiance. Rows (or blocks) in the highlight and shadow regions of the HDR scene could be overexposed and underexposed in the sRGB image. It would be very difficult or even impossible to accurately determine the optimal exposure times of these rows (or blocks). In addition, the optimal exposure time of every row (or block) is determined independently. Rowing (or blocking) and BOR artifacts could appear in the captured image.  Thus, it is  desired to develop a neural spatially varying AE algorithm with the temporary image as a raw image for HDR imaging. Clearly, each raw image captured by such a single-shot HDR imaging is different from a raw image captured by an existing CMOS sensor. It is also necessary to develop a new ISP for the single-shot HDR imaging. Therefore, future single-shot HDR pipelines should jointly consider exposure control, raw reconstruction, demosaicing, denoising, color correction, and tone mapping. In particular, row-wise or block-wise exposure patterns may introduce local exposure discontinuities and boundary artifacts, so reconstruction networks should explicitly model spatial exposure transitions and preserve consistent luminance relationships across neighboring regions. The conventional filter-based MEF algorithms in \cite{mertens2007exposure,1lizg2017,1kou2017} could be adopted to determine the ground-truth exposure times for all rows (or blocks) that are required for a supervised-learning-based row-wise or block-wise auto-exposure.

\begin{figure}[H]
\includegraphics[width=0.88\linewidth]{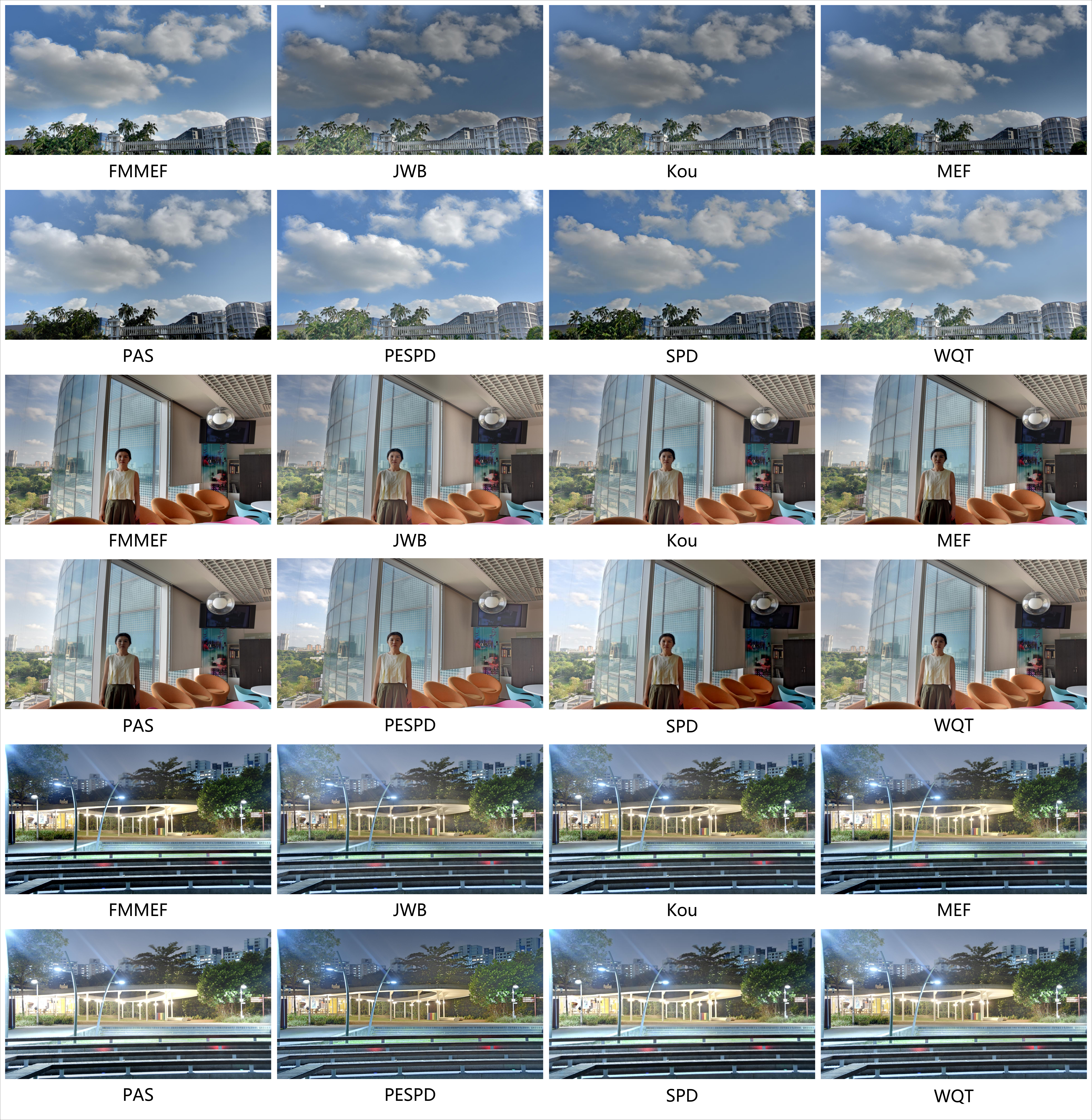}
\caption{Comparisons on real-world captured images with pixel-space methods.}
\label{fig:Brightness3}
\end{figure}

\begin{figure}[H]
\includegraphics[width=0.88\linewidth]{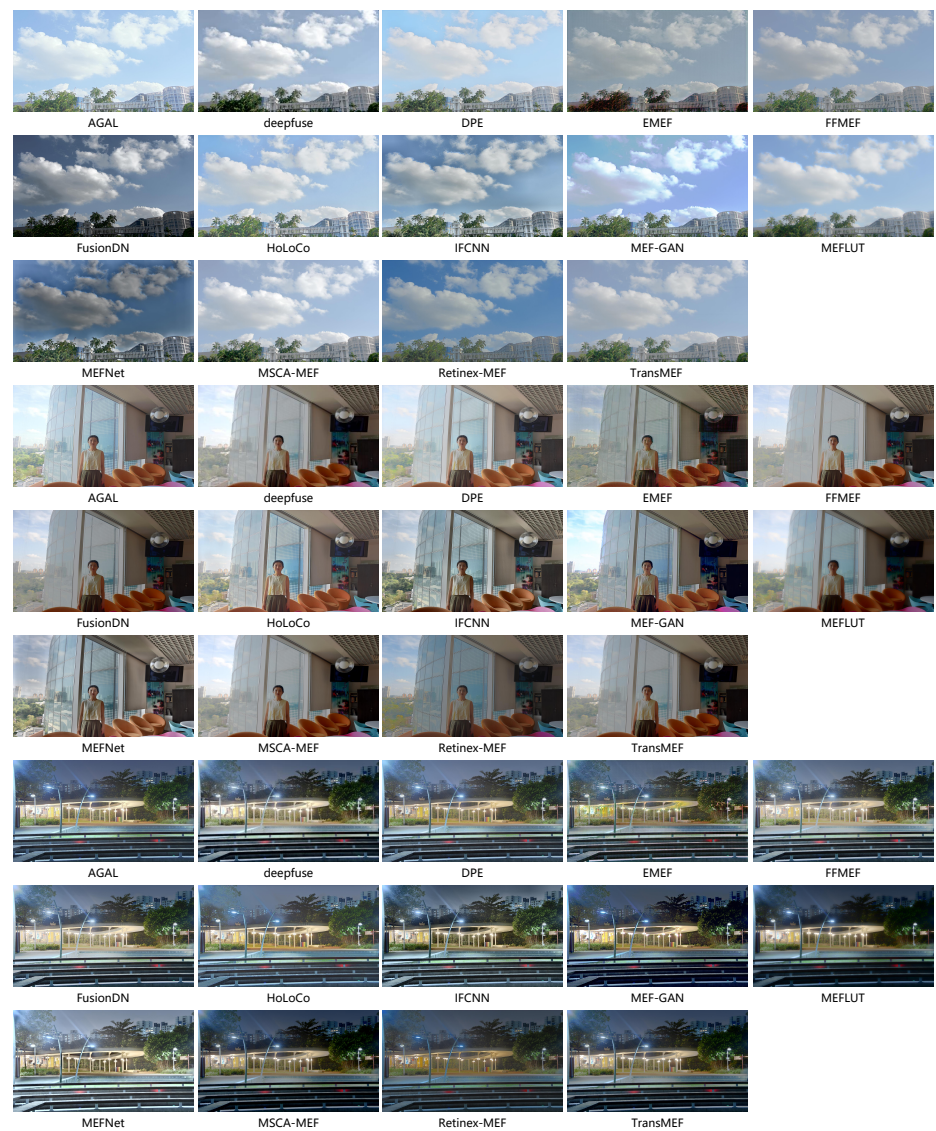}
\caption{Comparisons on real-world captured images with feature-space methods.}
\label{fig:Brightness4}
\end{figure}

Two large-exposure-ratio (LER) images can be captured in a single shot \cite{1tanaka2018,1tocci2011,zhang2021benchmarking}. The ratio should be as large as possible to capture more information from a real-world HDR scene. The two LER images can be  directly fused under the existing guideline, i.e., the fused image approaches the two LER images. Unfortunately,  there are BOR artifacts as shown in {Figure} \ref{fig:Brightness3}, and details in highlight/shadow regions of the real-world scene are not well preserved as demonstrated in Figure \ref{fig:Brightness4}. A new R\&D problem on MEF is "How to preserve scene depth and fine details of the real-world HDR scene in the fused image $Z_F$ with neither halo nor BOR artifacts even though the details are not captured by the set $\Omega_f$?".
Exposure interpolation \cite{li2023neural} and exposure extrapolation \cite{zheng2023neural} are helpful for solving this new problem. They could be addressed explicitly and implicitly. Explicit exposure interpolation and exposure extrapolation provide useful contextual information to guide implicit exposure interpolation and exposure extrapolation. 

One more interesting topic is low-exposure imaging. Motion blurring is an issue when the camera is with a fast-moving agent such as an autonomous vehicle (or a robot) or when objects to be captured move fast. Motion blur artifacts are reduced or removed if the exposure time $\Delta t_i$ is short.  Such a capturing method is called low-exposure imaging because $\Delta t_i/g_i$ is small. Low-exposure imaging is also very attractive in low-lighting conditions. The exposure time and sensor gain are widely adjusted to capture an image in low-lighting conditions. A clean but blurred image can be captured with a long $\Delta t_i$ and a large $g_i$  \cite{1zhangl2010}. A sharp but noisy image can be captured with an alternative setting of a short $\Delta t_i$ and a small $g_i$. These two settings also have a limitation to capture HDR scenes \cite{Debevec97} due to  washing-out artifacts \cite{1Hasinoffsw2010}. An attractive method for capturing an image in low-lighting conditions is to use low-exposure imaging. $\Delta t_i$ is maximized without introducing blurring artifacts while $g_i$ is minimized without causing overexposure. A sharp and clean image is captured.  Furthermore, the washing-out artifacts in the highlight regions of the captured image are significantly reduced due to  the small $\Delta t_i/g_i$. On the other hand, it is demonstrated from imaging models (\ref{imagingmodel}) and (\ref{imagingmodel2}) that the image $I_i$ is dark and  details in the darkest regions of the scene are not captured well. In addition, it can be shown from Equation (\ref{SNR}) that the SNR of the captured image could be low. It is thus desired to brighten the captured image without amplifying the noise and washing-out highlighted regions. 
The key challenge of low-exposure imaging is the trade-off among motion sharpness, signal-to-noise ratio, shadow detail recovery, and highlight preservation. Future methods should not simply brighten the low-exposure image but should jointly perform noise suppression, detail recovery, exposure compensation, and so on, preferably in the raw domain where physical exposure information is better preserved. In addition to optical imaging, low-exposure imaging is also useful for 3D X-ray imaging and microscope imaging. In fact, single-shot HDR imaging is also useful for them.

\section{Conclusions}

In this paper, we conducted a literature review on two important topics on high-dynamic-range (HDR) imaging: multi-exposure fusion (MEF) and ghost removal. We pointed out that ghost removal is still an open problem for multi-shot HDR imaging. Thus, single-shot HDR imaging is highly demanded. We provided a few interesting works on single-shot HDR imaging for future R\& D.

\vspace{+6pt}
\authorcontributions{Conceptualization, Q.T. and W.W.; methodology, Q.T.; software, Q.T.; validation, Q.T., W.W. and C.Z.; formal analysis, Q.T. and W.W.; investigation, Q.T.; resources, W.W.; data curation, Q.T.; writing---original draft preparation, Q.T.; writing---review and editing, W.W. and Z.L.; visualization, Q.T.; supervision, W.W. and Z.L.; project administration, W.W.; funding acquisition, W.W. All authors have read and agreed to the published version of the manuscript.}

\funding{This work was supported by the Natural Science Foundation of China (62202347).}

\institutionalreview{Not applicable.} 

\informedconsent{Not applicable.} 

\institutionalreview{Not applicable.}

\dataavailability{The data supporting the findings of this study are available within the article and its cited references.}

\conflictsofinterest{The authors declare no conflicts of interest.}

\begin{adjustwidth}{-\extralength}{0cm}

\reftitle{References}






\PublishersNote{}
\end{adjustwidth}
\end{document}